\documentclass[journal]{IEEEtran}
\usepackage{url}
\usepackage{algorithm}
\usepackage{algorithmic}
\usepackage{amssymb}

\newenvironment{proof}{{\indent \it Proof:\quad}}{$\hfill \blacksquare$}

\usepackage{xcolor}         
\usepackage{subfig}
\usepackage{multirow}
\usepackage{amsmath}
\usepackage{graphicx}
\usepackage{cite} 
\usepackage{makecell} 
\usepackage[colorlinks,
            linkcolor=blue,
            anchorcolor=blue,
            citecolor=blue,
            breaklinks]{hyperref}
\expandafter\def\expandafter\UrlBreaks\expandafter{%
  \UrlBreaks
  \do\a\do\b\do\c\do\d\do\e\do\f\do\g\do\h\do\i\do\j%
  \do\k\do\l\do\m\do\n\do\o\do\p\do\q\do\r\do\s\do\t%
  \do\u\do\v\do\w\do\x\do\y\do\z%
  \do\A\do\B\do\C\do\D\do\E\do\F\do\G\do\H\do\I\do\J%
  \do\K\do\L\do\M\do\N\do\O\do\P\do\Q\do\R\do\S\do\T%
  \do\U\do\V\do\W\do\X\do\Y\do\Z%
  \do\0\do\1\do\2\do\3\do\4\do\5\do\6\do\7\do\8\do\9}

\usepackage{cleveref}

\ifCLASSINFOpdf
\else
\fi
\begin{document}
%
\title{Information-Geometric Inverse Distillation for Enhancing Adversarial Transferability}
%
%
%


\author{Wenyuan~Wu,
        Yuan~Sun,
        Yingke~Chen,
        Chao~Su,
        Xi~Peng,
        Dezhong~Peng,
        Xu~Wang%
\thanks{\textit{Corresponding author: Xu Wang.}}%
\thanks{Wenyuan~Wu, Yuan~Sun, Chao~Su, Dezhong~Peng, Xu~Wang are with the College of Computer Science, Sichuan University, China. Xi~Peng is with the School of Artificial Intelligence,  Sichuan University, China. (email: wuwenyuan97@gmail.com; sunyuan\_work@163.com; suchao.ml@gmail.com; pengx.gm@gmail.com; pengdz@scu.edu.cn; wangxu.scu@gmail.com).}%
\thanks{Yingke~Chen is with the Department of Computer and Information Sciences, Northumbria University, UK. (email: yingke.chen@northumbria.ac.uk).}}

%
%

\markboth{Journal of \LaTeX\ Class Files,~Vol.~14, No.~8, August~2015}%
{Shell \MakeLowercase{\textit{et al.}}, Bare Demo of IEEEtran.cls for IEEE Journals}
%



\maketitle

\begin{abstract}
Transfer-based adversarial attacks rely on surrogate models to craft perturbations, yet often overfit the surrogate's decision boundary. To address this problem, we propose Inverse Knowledge Distillation (IKD), a simple and attack-agnostic mechanism that maximizes the prediction-distribution discrepancy between benign and adversarial samples on the surrogate model. IKD uses a CE/KL-equivalent soft-label objective to push adversarial predictions away from a fixed benign prediction anchor and enrich the attack with Fisher-sensitive surrogate directions. We prove that, under a matched fixed-anchor implementation, soft-label cross-entropy and KL divergence differ only by a constant entropy term and therefore induce identical gradients, Hessians, and adversarial optimization trajectories. Our information-geometric analysis further derives a quantitative lower bound on dominant Fisher-subspace overlap between surrogate and target models from local same-task stability and a Fisher eigengap, and establishes a sufficient target-margin crossing condition under oriented gradient coherence and target smoothness. This analysis connects IKD's surrogate Fisher sensitivity to cross-model transfer. In contrast, mean squared error uses a different Euclidean pullback in output probability space. IKD integrates seamlessly with standard gradient-based attacks without modifying their optimization pipelines. Extensive ImageNet experiments demonstrate consistent black-box gains across CNN, ViT, and defended models, while ablations confirm CE and KL equivalence and the pronounced disadvantage of MSE. These results establish IKD as an effective and lightweight component for improving adversarial transferability. Code is available at https://github.com/ImmortalTing/IKD.
\end{abstract}


%
\IEEEpeerreviewmaketitle

\section{Introduction}\label{Introduction}
\IEEEPARstart{D}{eep} neural networks (DNNs) achieve remarkable success in a wide range of applications, including facial recognition, autonomous driving, and medical diagnosis. However, a growing body of work shows that DNNs are highly vulnerable to adversarial examples, which are inputs perturbed by imperceptible noise that can drastically change the model prediction~\cite{szegedy2013intriguing}. This vulnerability raises serious security concerns, as it exposes safety-critical systems to deliberate attacks in realistic deployment scenarios. While a vast literature focuses on hardening models against adversarial perturbations, understanding and improving attack mechanisms remains equally important for stress-testing deployed systems and uncovering fundamental weaknesses of deep models.

Most existing attacks generate adversarial examples by adding carefully crafted perturbations to benign inputs. These perturbations are either synthesized by generative networks~\cite{zhao2018generating} or, more commonly, obtained via gradient-based optimization~\cite{goodfellow2014explaining,madry2017towards,dong2018boosting,xie2019improving,dong2019evading,lin2019nesterov}. In gradient-based attacks, the perturbation is iteratively updated along the gradient of a task-specific loss, such as cross-entropy, with respect to the input. Intuitively, this procedure searches for directions in the input space that can most effectively increase the loss on a given model.


\begin{figure*}[tbp]
    \centering
    \includegraphics[width=1.0\textwidth]{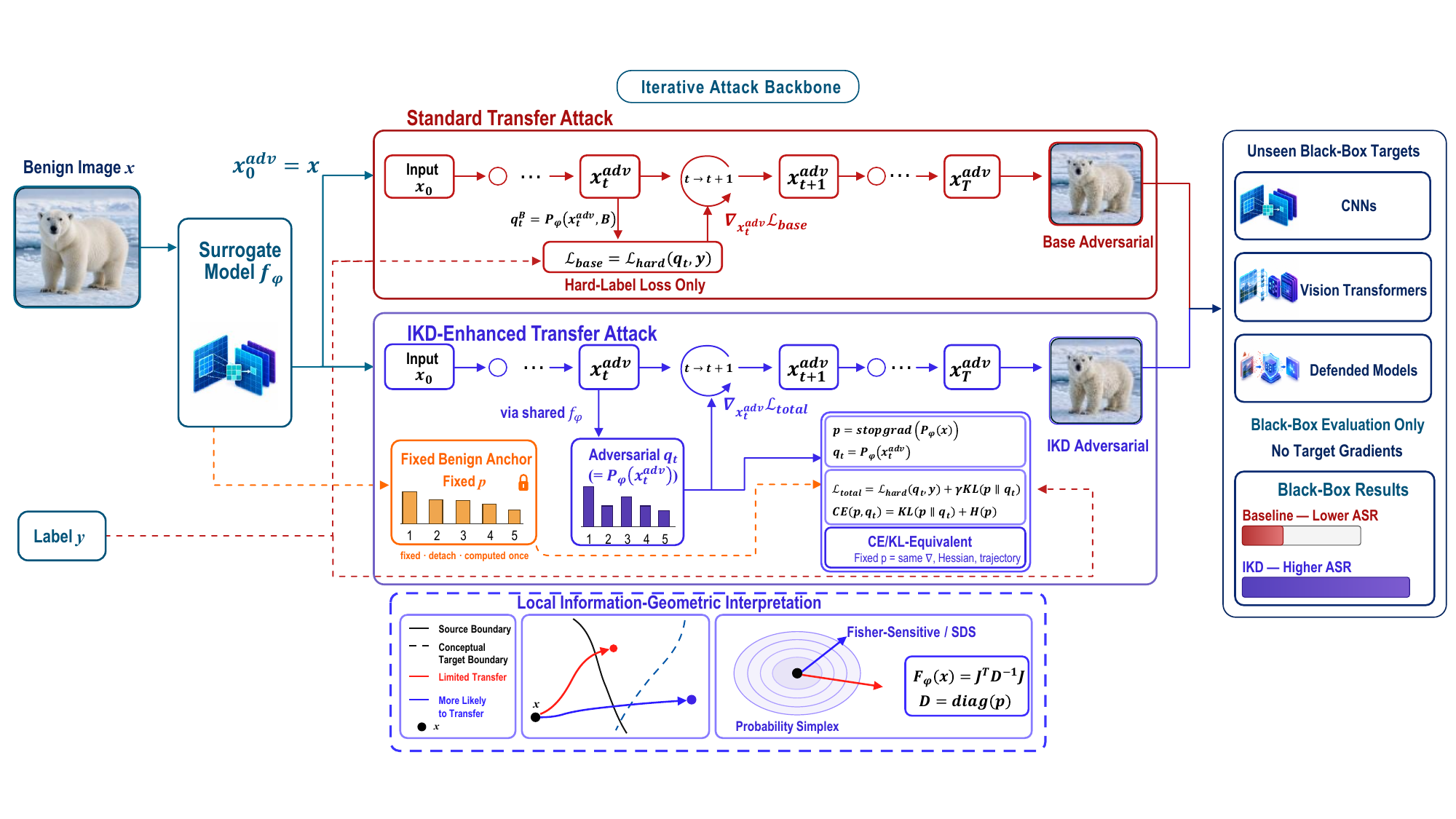}
    \caption{\textbf{Standard and IKD-enhanced transfer attacks.} Both branches share the input, label, fixed surrogate, initialization, and attack backbone but maintain independent states. IKD augments the hard-label loss with $\gamma\,\mathrm{KL}(p\|q_t)$, where $p=\operatorname{stopgrad}(f_\varphi(x))$ is fixed and $q_t=f_\varphi(x_t^{adv})$. Soft-label CE is equivalent under matched implementations. Both outputs are evaluated on unseen black-box targets. The lower panel illustrates how Fisher-sensitive directions support transfer through the subspace-sharing and target-margin conditions developed in Section~\ref{sec:cross-model-sharing}.}
    \label{fig:structur}
\end{figure*}

Although gradient-based attacks are highly effective in the white-box setting, they become much less practical for black-box models where internal information is inaccessible. In such cases, an attacker typically resorts to either query-based or transfer-based strategies. Query-based attacks estimate gradients through repeated queries to the target model, which incurs substantial computational and time costs. Transfer-based attacks, in contrast, generate adversarial examples on one or more surrogate models and exploit their transferability to attack the black-box model. Owing to their efficiency and practicality, transfer-based attacks attract significant attention and are the focus of this paper.

A rich line of transfer-based attacks includes input transformation-based methods~\cite{dong2019evading,liang2023styless,wang2021admix,xie2019improving,wang2024boosting}, advanced gradient-based updates~\cite{dong2018boosting,li2023adaptive,lin2019nesterov,wang2021enhancing,niu2025enhancing}, ensemble-based strategies~\cite{qian2023lea2,tramer2018ensemble,xiong2022stochastic}, and feature-space attacks~\cite{ganeshan2019fda,wang2023improving,zhang2022improving}, among others. While these methods can improve transferability to some extent, they largely operate at the level of heuristics (e.g., input diversity, momentum, gradient smoothing) and typically require nontrivial additional computation. More importantly, most existing approaches still optimize the loss of a particular surrogate model and therefore remain implicitly tied to its decision boundary. As a consequence, the resulting adversarial perturbations tend to overfit the surrogate. They achieve high success rates on the surrogate model but do not generalize equally well to unseen target models. From a geometric viewpoint, the attack is driven by gradients determined by the surrogate, while other transfer-relevant surrogate-sensitive directions remain unexplored.

To address this limitation, we revisit transfer-based attacks from a knowledge distillation perspective and explicitly focus on the mismatch between surrogate and target models. We propose Inverse Knowledge Distillation (IKD)\footnote{Here, ``distillation'' refers to maximizing a prediction-distribution discrepancy between an adversarial prediction and a fixed, detached benign prediction anchor on the same surrogate model, rather than conventional knowledge distillation.}, a simple yet effective mechanism that enhances gradient diversity and mitigates surrogate overfitting. As illustrated in~\Cref{fig:structur}, the standard and IKD-enhanced attacks use the same benign input $x$, label $y$, fixed-parameter surrogate $f_\varphi$, initialization $x_0^{adv}=x$, and iterative attack backbone, but evolve along independent attack states. The standard branch maximizes only the hard-label objective. IKD additionally computes $p=\operatorname{stopgrad}(f_\varphi(x))$ once before the attack, recomputes $q_t=f_\varphi(x_t^{adv})$ at each iteration, and maximizes $\mathcal{L}_{hard}(q_t,y)+\gamma\,\mathrm{KL}(p\|q_t)$. Under matched fixed-anchor implementation conditions, soft-label CE provides an optimization-equivalent realization of the same soft objective. The attacks are generated exclusively from surrogate information and then evaluated on unseen black-box targets. By enlarging this prediction-distribution discrepancy, IKD explores additional Fisher-sensitive directions and reduces dependence on the surrogate's specific decision boundary. Section~\ref{sec:cross-model-sharing} establishes how local same-task stability and a Fisher eigengap produce dominant Fisher-subspace overlap between surrogate and target models, and how oriented coherence and target smoothness convert this shared sensitivity into target-margin progress.

The main contributions of this work are summarized below.
\begin{itemize}
    \item We introduce a knowledge-distillation-based view of transfer-based adversarial attacks and instantiate it with a simple and effective Inverse Knowledge Distillation (IKD) attack. By explicitly enlarging the prediction-distribution discrepancy between benign and adversarial inputs on the surrogate model, IKD mitigates surrogate overfitting and substantially improves transferability across heterogeneous models. The proposed IKD term is model-agnostic, incurs negligible additional overhead, and can be seamlessly integrated into a wide range of gradient-based attacks.
    \item We systematically investigate different soft-label discrepancy objectives for IKD and reveal that their optimization behavior is determined by the geometry they induce in the output probability space. In particular, we show that soft-label cross-entropy (CE) and Kullback--Leibler (KL) divergence are optimization-equivalent under matched fixed-anchor implementation conditions. They differ only by the constant entropy term $H(p)$, and therefore lead to identical gradients, Hessians, iterative attack trajectories, and experimental results. In contrast, mean squared error (MSE) uses an isotropic Euclidean output-space metric whose input-space pullback differs from the Fisher pullback, explaining the substantial MSE degradation observed across the attack suite.
    \item Beyond empirical gains, we provide an information-geometric analysis of IKD. We first show that the CE/KL-equivalent objective locally emphasizes principal directions of the surrogate Fisher geometry. We then prove that local same-task stability relative to a Fisher eigengap yields a quantitative bound on dominant Fisher-subspace overlap between surrogate and target models, and that oriented gradient coherence and target smoothness yield a sufficient target-margin crossing condition. Together, these results establish an explicit theoretical connection between shared Fisher-sensitive structure and adversarial transfer.
    \item We conduct extensive experiments on the ImageNet dataset with diverse convolutional and transformer-based architectures, including robust and defended models. Across diverse attack backbones and pairs of surrogate and target models, IKD substantially improves transfer success rates, including cross-architecture and defended targets. Ablation studies further confirm CE and KL equivalence and reveal the pronounced performance disadvantage of MSE. These results demonstrate the practical effectiveness and broad applicability of IKD.
\end{itemize}

\section{Related Works}
\subsection{Adversarial Attack}
Adversarial attacks use generative approaches~\cite{zhao2018generating} and gradient-based optimization~\cite{goodfellow2014explaining,madry2017towards,dong2018boosting}. Transferability is enhanced through input transformations~\cite{xie2019improving,dong2019evading,lin2019nesterov} and gradient refinement based on momentum~\cite{nesterov1983method}, variance tuning~\cite{wang2021enhancing}, or enhanced momentum~\cite{wang2021boosting}.

Other strategies use adaptive perturbation updates~\cite{yuan2024adaptive}.

Zheng et al.~\cite{zheng2025multifeature} improve transferability by aggregating attention across multiple feature layers and disrupting the resulting ensemble attention. Wang et al.~\cite{wang2024frequencyaware} concentrate perturbation optimization on selected frequency components that contribute strongly to model predictions. In contrast, IKD augments a conventional hard-label attack loss with a discrepancy between benign and adversarial prediction distributions. Its information-geometric analysis connects the Fisher-sensitive surrogate directions emphasized by this objective to cross-model transfer.

\subsection{Geometric Accounts of Adversarial Transfer}
Prior work provides several complementary, but non-equivalent, geometric views of transfer. Input-space Fisher geometry is used to identify locally sensitive adversarial directions of a single model~\cite{zhao2019fisher}. Separately, Tram\`er et al.~\cite{tramer2017space} empirically identify shared adversarial subspaces, while Charles et al.~\cite{charles2019geometric} establish common adversarial directions for restricted classifier families. Analyses based on loss gradients further associate transfer with alignment of source and target gradients~\cite{demontis2019why} and show that gradient similarity and model smoothness jointly affect transfer bounds~\cite{yang2021trs}. Together, these studies motivate a unified connection between Fisher-sensitive directions, shared adversarial structure, and transfer. Section~\ref{sec:cross-model-sharing} develops this connection by proving that local same-task predictor stability and spectral separation yield dominant Fisher-subspace overlap and by linking this overlap to a sufficient target-margin condition.

\subsection{Adversarial Defense}

Adversarial defenses primarily detect suspicious inputs~\cite{meng2017magnet,liang2018detecting}, purify adversarial images before classification~\cite{liao2018defense,liu2019feature,jia2019comdefend}, or improve model robustness through adversarial training~\cite{madry2017towards,tramer2018ensemble,pang2020boosting}.

Prediction-distribution discrepancies are also explored in robustness-oriented learning. Virtual Adversarial Training (VAT)~\cite{miyato2018virtual} seeks local perturbations that maximize the divergence between predictions on an input and its perturbed counterpart. It then penalizes this divergence during training to promote local smoothness. TRADES~\cite{zhang2019theoretically} combines a natural classification loss with a weighted worst-case prediction discrepancy. This objective balances natural accuracy and adversarial robustness. In both methods, discrepancy maximization identifies local variations that subsequent model updates aim to suppress.

In contrast, IKD retains discrepancy maximization as an attack objective and keeps the surrogate fixed throughout. The optimized inputs are the final adversarial examples, rather than intermediate samples used to regularize the model. IKD couples the soft prediction-discrepancy term with a conventional hard-label attack loss to improve transferability to unseen black-box targets. This formulation gives prediction discrepancy a direct role in attack generation through inverse distillation. The accompanying information-geometric analysis links the Fisher-sensitive surrogate directions emphasized by this objective to cross-model adversarial transfer.

\section{Methodology}\label{Method}

\subsection{Problem Setup}\label{section:Problem Definition}
Given a benign image $x$ with ground-truth label $y$, the goal of an adversarial attack is to generate a perturbed input $x^{adv}=x+\delta$ that causes misclassification while remaining visually indistinguishable from $x$. A standard gradient-based formulation solves
\begin{equation}\label{equation:optimal advsample}
    \begin{aligned}
        \mathop{\arg\max}\limits_{x^{adv}}~&\mathcal{L}(f_\theta(x^{adv}),y) \\
        \text{s.t.}~&\|x^{adv}-x\|_p \le \epsilon,
    \end{aligned}
\end{equation}
where $f_\theta$ is a trained classifier, $\mathcal{L}$ is the cross-entropy loss, and $\|\cdot\|_p$ denotes the $L_p$ norm. Following prior work~\cite{dong2019evading,li2023adaptive,xie2019improving}, we adopt the $L_\infty$ constraint, i.e., $\|\delta\|_\infty \le \epsilon$, which restricts each pixel perturbation to lie within a small range and keeps the adversarial example perceptually close to the benign input.

The optimization in~\eqref{equation:optimal advsample} assumes white-box access to $f_\theta$. In a black-box setting, the target model is inaccessible, and directly maximizing $\mathcal{L}(f_\theta(x^{adv}),y)$ is infeasible. A common strategy is to introduce a surrogate model $f_\varphi$, generate adversarial examples on $f_\varphi$, and exploit their transferability to attack the inaccessible target $f_\theta$. Since $f_\varphi$ and $f_\theta$ are trained independently and may differ substantially in architecture and parameters, improving the transferability of perturbations crafted on $f_\varphi$ is critical. In this work, we design attack objectives that reduce surrogate overfitting and yield perturbations that transfer reliably to unseen black-box models.

\subsection{Inverse Knowledge Distillation Objective}\label{section:Inverse Knowledge Distillation Attack}
Knowledge distillation (KD) is a model compression technique that transfers the ``dark knowledge'' of a large teacher network to a smaller student by matching either its output probability distribution or intermediate representations~\cite{hinton2015distilling}. In standard KD, the student is trained to approximate the teacher's predictions, thereby preserving its behavior with fewer parameters.

Inspired by this idea, we propose Inverse Knowledge Distillation (IKD) for transfer-based attacks. Instead of aligning two models, IKD operates on a single surrogate model $f_\varphi$ and enforces its predictive-probability distribution on adversarial inputs to be maximally different from that on benign inputs. That is, rather than distilling knowledge from a teacher to a student, we deliberately “distill away” the surrogate-specific prediction anchor embedded in $f_\varphi(x)$ when crafting $x^{adv}$.

Formally, given a surrogate $f_\varphi$, we modify the standard attack objective~\eqref{equation:optimal advsample} to
\begin{equation}\label{equation:IKD optimal}
    \begin{aligned}
        \mathop{\arg\max}\limits_{x^{adv}}~&
        \mathcal{L}_{hard}(f_\varphi(x^{adv}),y) 
        + \gamma\,\mathcal{L}_{soft}(f_\varphi(x^{adv}),f_\varphi(x)) \\
        \text{s.t.}~&\|x^{adv}-x\|_\infty \le \epsilon,
    \end{aligned}
\end{equation}
where $p=f_\varphi(x)$ and $q=f_\varphi(x^{adv})$ are predictive probabilities. Here, $p$ is the benign prediction anchor, detached from the optimization graph and fixed during adversarial optimization, while $q$ is the adversarial prediction. $\mathcal{L}_{hard}$ is the standard classification loss, $\mathcal{L}_{soft}$ is a soft-label prediction-distribution discrepancy between $p$ and $q$, and $\gamma>0$ controls the trade-off. Intuitively, $\mathcal{L}_{hard}$ ensures that $x^{adv}$ misleads the surrogate, while $\mathcal{L}_{soft}$ explicitly pushes the adversarial prediction distribution $q$ away from the benign prediction distribution $p$. Together, the hard- and soft-label terms broaden the surrogate-local directions explored by the attack. Section~\ref{sec:cross-model-sharing} shows how their dominant Fisher components overlap with target-sensitive directions and drive transferred target-margin changes. Existing input transformations and gradient smoothing alter the sample or update, whereas IKD changes the output-distribution objective.

In principle, the soft-label loss $\mathcal{L}_{soft}$ in~\eqref{equation:IKD optimal} can be instantiated by various discrepancies between the two output distributions, such as mean squared error (MSE), soft-label cross-entropy (CE), or KL divergence. We adopt KL as the canonical form because it directly exposes the information geometry underlying IKD. The corresponding loss is
\begin{equation}\label{eq:ikd-kl}
    \begin{aligned}
        \mathcal{L}_{soft}(f_\varphi(x^{adv}),f_\varphi(x))
        &= \mathrm{KL}(f_\varphi(x)\,\|\, f_\varphi(x^{adv})) \\
        &= \sum_i f_\varphi(x)(i)\,\log\frac{f_\varphi(x)(i)}{f_\varphi(x^{adv})(i)}.
    \end{aligned}
\end{equation}

Equivalently, one may instantiate the same IKD objective using soft-label cross-entropy
\begin{equation}\label{eq:ikd-ce}
    \mathcal{L}_{soft}^{\mathrm{CE}}(q,p)=\mathrm{CE}(p,q)=-\sum_i p_i\log q_i.
\end{equation}

Since the benign anchor $p=f_\varphi(x)$ is fixed during adversarial optimization, CE and KL satisfy
\begin{equation}\label{eq:ikd-ce-kl}
    \mathrm{CE}(p,q)=\mathrm{KL}(p\|q)+H(p),
\end{equation}
where $H(p)=-\sum_i p_i\log p_i$ is independent of $x^{adv}$. Therefore, CE and KL are optimization-equivalent in the proposed fixed-anchor IKD formulation.

This equivalence can also be seen from the first-order gradients. For both KL and soft-label CE, the derivative with respect to the adversarial prediction component $q_i$ is
\begin{equation}\label{eq:ikd-prediction-gradient}
    \frac{\partial \mathrm{KL}(p\|q)}{\partial q_i} = \frac{\partial \mathrm{CE}(p,q)}{\partial q_i} = -\frac{p_i}{q_i}.
\end{equation}
By the chain rule, they induce the same gradient with respect to $x^{adv}$
\begin{equation}\label{eq:ikd-input-gradient}
    \nabla_{x^{adv}} \mathrm{KL}(p\|q) = \nabla_{x^{adv}} \mathrm{CE}(p,q).
\end{equation}

Under sign-gradient or normalized-gradient iterative attacks, matching the fixed and detached anchor $p$, temperature, reduction, soft-loss weight $\gamma$, and numerical implementation preserves identical CE/KL update directions and adversarial trajectories. When the CE and KL reductions differ by a constant scaling, rescaling $\gamma$ preserves this equivalence.

\subsection{Theoretical Analysis of IKD}\label{sec:info_geom}
\begin{figure*}[htbp]
    \centering
    \includegraphics[width=1\textwidth]{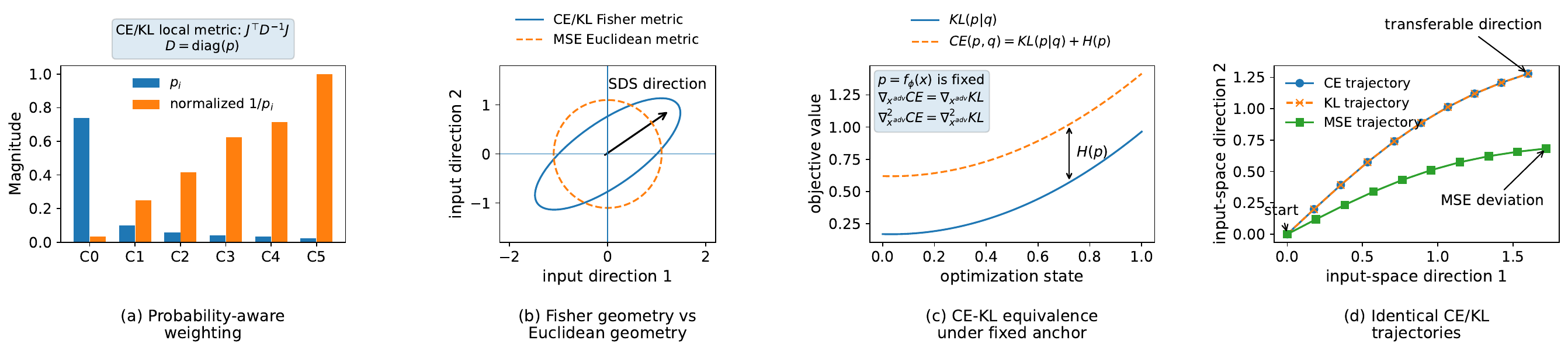}
    \caption{
    \textbf{Information-geometric analysis of IKD.}
    (a) The CE/KL Fisher pullback $J^\top D^{-1}J$, with $D=\mathrm{diag}(p)$, weights output dimensions by inverse probability.
    (b) Fisher-sensitive directions support cross-model transfer under the conditions in Section~\ref{sec:cross-model-sharing}, whereas MSE induces the Euclidean pullback $J^\top J$.
    (c) For fixed $p$, $\mathrm{CE}(p,q)=\mathrm{KL}(p\|q)+H(p)$ differs only by a constant.
    (d) CE and KL therefore yield overlapping attack trajectories under matched implementations, while MSE follows different geometry. Panels illustrate the theoretical relationships.
    }
    \label{fig:proof}
\end{figure*}

We present a unified theoretical analysis of inverse knowledge distillation (IKD), visually summarized in~\Cref{fig:proof}, from the perspective of output-distribution geometry. The analysis establishes four results. These are (1) exact optimization equivalence between soft-label cross-entropy (CE) and KL divergence under the fixed benign prediction anchor; (2) the local pullback Fisher geometry induced by the CE/KL-equivalent IKD objective; (3) a quantitative lower bound on cross-model Fisher-subspace overlap and a sufficient local transfer guarantee; and (4) a geometric explanation for the performance degradation caused by MSE.

\subsubsection{Equivalence of CE and KL under Fixed Soft-Label Anchors}


Let
\begin{equation}
    p = f_\varphi(x), \quad q = f_\varphi(x^{adv}),
\end{equation}
where $p$ is the benign prediction anchor and $q$ is the adversarial prediction. During adversarial optimization, $p$ is fixed and detached from the optimization graph, while $q$ changes with $x^{adv}$.


The soft-label CE and KL objectives are defined in~\Cref{eq:ikd-ce,eq:ikd-kl}, and their constant-offset identity is given in~\Cref{eq:ikd-ce-kl}. Here,
\begin{equation}\label{eq:anchor-entropy}
    H(p)=-\sum_i p_i\log p_i.
\end{equation}
This constant-offset relationship is illustrated in~\Cref{fig:proof}(c), where CE and KL have the same shape and differ only by the anchor entropy \(H(p)\).




\textbf{Proposition 1. Optimization equivalence of CE and KL}. Under the fixed benign prediction anchor $p=f_\varphi(x)$, soft-label CE and KL divergence differ only by the entropy term $H(p)$, which is independent of $x^{adv}$. The shared input gradient is stated in~\Cref{eq:ikd-input-gradient}. They also induce identical Hessians with respect to the adversarial input, as shown below.
\begin{equation}\label{eq:second-order}
    \nabla^2_{x^{adv}} \mathrm{CE}(p,q)=\nabla^2_{x^{adv}} \mathrm{KL}(p\|q).
\end{equation}

\begin{proof}
    The identity in~\Cref{eq:ikd-ce-kl} gives the constant difference $H(p)$. Since $p=f_\varphi(x)$ is fixed during adversarial optimization, $H(p)$ does not depend on $x^{adv}$. Therefore,
    \begin{equation}
        \nabla_{x^{adv}}H(p)=0, \qquad \nabla^2_{x^{adv}}H(p)=0.
    \end{equation}
    Taking the first- and second-order derivatives of~\Cref{eq:ikd-ce-kl} with respect to $x^{adv}$ gives~\Cref{eq:ikd-input-gradient,eq:second-order}.
\end{proof}

This equivalence also appears at the prediction level through the shared derivative in~\Cref{eq:ikd-prediction-gradient}. By the chain rule, CE and KL therefore provide identical gradients with respect to $x^{adv}$. Under these matched conditions, iterative sign-gradient or normalized-gradient attacks produce the same update directions and the same adversarial trajectory. \Cref{fig:proof}(d) visualizes this consequence. The CE and KL trajectories completely overlap, whereas the MSE trajectory deviates because it follows a different gradient geometry.

\subsubsection{Fisher Geometry of the CE/KL-Equivalent IKD Objective}
We next analyze the local geometry induced by the CE/KL-equivalent objective. Assume that $f_\varphi:\mathbb{R}^d\rightarrow\operatorname{int}\Delta^{C-1}$ is twice continuously differentiable in a neighborhood of $x$. Denote the Jacobian and probability matrix as
\begin{equation}
    J=\frac{\partial f_\varphi(x)}{\partial x}, \qquad D=\mathrm{diag}(p).
\end{equation}
We consider $x^{adv}$ in a neighborhood of $x$, define $\delta=x^{adv}-x$, and let $\Delta=q-p$. Then
\begin{equation}
    \Delta=J\delta+O(\|\delta\|_2^2), \qquad \mathbf{1}^\top\Delta=0,
\end{equation}
where the second identity follows because both $p$ and $q$ lie on the probability simplex.


\textbf{Proposition 2. Second-order expansion of the CE/KL-equivalent objective}. Let $p_{\min}=\min_i p_i>0$. As $\|\delta\|_2\rightarrow0$, KL satisfies
\begin{equation}\label{eq:kl-taylor}
    \mathrm{KL}(p\|q)=\frac{1}{2}\delta^\top J^\top D^{-1}J\delta+O(\|\delta\|^3),
\end{equation}
and CE satisfies
\begin{equation}\label{eq:CE-taylor}
    \mathrm{CE}(p,q)=H(p)+\frac{1}{2}\delta^\top J^\top D^{-1}J\delta+O(\|\delta\|^3).
\end{equation}

\begin{proof}
    Since $q=p+\Delta$, expanding the logarithm gives
    \begin{equation}
        \log\frac{p_i}{p_i+\Delta_i} =-\log\left(1+\frac{\Delta_i}{p_i}\right) =-\frac{\Delta_i}{p_i}+\frac{1}{2}\frac{\Delta_i^2}{p_i^2}+O(\Delta_i^3).
    \end{equation}
    Thus,
    \begin{equation}
        \mathrm{KL}(p\|q)=-\sum_i\Delta_i+\frac{1}{2}\sum_i\frac{\Delta_i^2}{p_i}+O(\|\Delta\|^3).
    \end{equation}
    Since $\sum_i\Delta_i=0$,
    \begin{equation}
        \mathrm{KL}(p\|q)=\frac{1}{2}\sum_i\frac{\Delta_i^2}{p_i}+O(\|\Delta\|^3).
    \end{equation}
    Substituting $\Delta=J\delta+O(\|\delta\|_2^2)$ yields~\Cref{eq:kl-taylor}; combining it with~\Cref{eq:ikd-ce-kl} gives~\Cref{eq:CE-taylor}.
\end{proof}



The matrix $F_\varphi(x)=J^\top D^{-1}J$ is the local pullback of the categorical Fisher metric to the input space~\cite{amari2016information,zhao2019fisher}. If $g_{\varphi,i}=\nabla_x\log p_i$ denotes the class-$i$ score gradient, then
\begin{equation}\label{eq:fisher-score-identity}
    F_\varphi(x)=\sum_{i=1}^{C}p_i g_{\varphi,i}g_{\varphi,i}^{\top},
    \quad
    v^\top F_\varphi(x)v=\sum_{i=1}^{C}p_i\langle g_{\varphi,i},v\rangle^2.
\end{equation}
Thus, a large Fisher quadratic form quantifies high expected local sensitivity of the surrogate predictive distribution. The analysis below augments this sensitivity with orientation and margin information to derive a target-misclassification condition. Through the eigendecomposition
\begin{equation}\label{eq:fisher-eigendecomposition}
    J^\top D^{-1}J=U\Lambda U^\top,
\end{equation}
where $U$ is orthogonal and $\lambda_1\geq\cdots\geq\lambda_d\geq0$, we define the top-$k$ \emph{surrogate Fisher-sensitive subspace} as
\begin{equation}
    \mathcal{S}_\varphi^{(k)}=\mathrm{span}\{u_1,\ldots,u_k\}.
\end{equation}
Under the CE/KL-equivalent objective,
\begin{equation}
    \delta^\top J^\top D^{-1}J\delta=\sum_{i=1}^{d}\lambda_i\langle \delta,u_i\rangle^2.
\end{equation}
Consequently, the local quadratic term assigns larger weight to perturbation components along high-eigenvalue surrogate directions. The next subsection extends this surrogate Fisher geometry to cross-model subspace sharing through predictor stability and spectral separation.

\subsubsection{Cross-Model Fisher-Subspace Sharing and Transfer}\label{sec:cross-model-sharing}
The target model enters the following analysis to establish cross-model transfer properties, while attack generation remains entirely surrogate-based. For $m\in\{s,t\}$, let $f_m:\mathbb{R}^d\rightarrow\operatorname{int}\Delta^{C-1}$ denote the differentiable predictive map of the surrogate or target, with common input coordinates and class indices, and define
\begin{equation}
\begin{aligned}
    p_m&=f_m(x), & J_m&=\frac{\partial f_m(x)}{\partial x},\\
    D_m&=\mathrm{diag}(p_m), & F_m&=J_m^\top D_m^{-1}J_m.
\end{aligned}
\end{equation}

\textbf{Assumption 1 (local same-task stability).}
There exists a twice continuously differentiable reference task predictor $f_\star$, with $p_\star=f_\star(x)$, $D_\star=\mathrm{diag}(p_\star)$, $J_\star=\partial f_\star(x)/\partial x$, and $F_\star=J_\star^\top D_\star^{-1}J_\star$, such that for $m\in\{s,t\}$,
\begin{equation}\label{eq:task-stability}
\begin{aligned}
    &\min_i\{p_{m,i},p_{\star,i}\}\geq\nu>0,\qquad
      \|J_m\|_2,\|J_\star\|_2\leq M,\\
    &\|p_m-p_\star\|_\infty\leq\varepsilon_{p,m},\qquad
      \|J_m-J_\star\|_2\leq\varepsilon_{J,m}.
\end{aligned}
\end{equation}
Let $g_k=\lambda_k(F_\star)-\lambda_{k+1}(F_\star)>0$ and define
\begin{equation}\label{eq:fisher-perturbation-radius}
    \xi_m=\frac{2M\varepsilon_{J,m}}{\nu}
          +\frac{M^2\varepsilon_{p,m}}{\nu^2},
    \qquad \xi_m<\frac{g_k}{2}.
\end{equation}
This predictor-level formulation enables Fisher-subspace sharing to be derived directly from predictive probabilities and their input Jacobians.

\textbf{Proposition 3 (cross-model sharing of dominant Fisher subspaces).}
Let $P_m$ and $P_\star$ be the orthogonal projectors onto the top-$k$ eigenspaces of $F_m$ and $F_\star$, respectively. Under Assumption~1,
\begin{equation}\label{eq:projector-bound}
\begin{aligned}
    \|F_m-F_\star\|_2&\leq\xi_m,\\
    \|P_s-P_t\|_2&\leq r_k
    :=\frac{\xi_s}{g_k-\xi_s}
      +\frac{\xi_t}{g_k-\xi_t}.
\end{aligned}
\end{equation}
Moreover, with $U_{m,k}$ containing the top-$k$ eigenvectors, the normalized overlap
\begin{equation}\label{eq:fisher-overlap}
    \Omega_k:=\frac{1}{k}\|U_{s,k}^\top U_{t,k}\|_F^2
    =\frac{1}{k}\mathrm{tr}(P_sP_t)
    \geq 1-\min\{1,r_k\}^2.
\end{equation}
Hence, the quantitative local-stability bounds relative to an eigengap guarantee large overlap when $r_k\ll1$; in this regime, the dominant surrogate and target directions form an approximately shared discriminative subspace (SDS).

\begin{proof}
For $m\in\{s,t\}$, add and subtract terms to write
\begin{equation}
\begin{aligned}
F_m-F_\star={}&(J_m-J_\star)^\top D_m^{-1}J_m
 +J_\star^\top D_m^{-1}(J_m-J_\star)\\
&+J_\star^\top(D_m^{-1}-D_\star^{-1})J_\star.
\end{aligned}
\end{equation}
Equation~\eqref{eq:task-stability} gives $\|D_m^{-1}\|_2\leq1/\nu$ and $\|D_m^{-1}-D_\star^{-1}\|_2\leq\varepsilon_{p,m}/\nu^2$, proving the first inequality in~\eqref{eq:projector-bound}. Weyl's inequality gives the separation $\lambda_k(F_\star)-\lambda_{k+1}(F_m)\geq g_k-\xi_m$. The Davis--Kahan $\sin\Theta$ theorem~\cite{yu2015davis} therefore gives
\begin{equation}
    \|P_m-P_\star\|_2\leq\frac{\xi_m}{g_k-\xi_m}.
\end{equation}
The triangle inequality proves the projector bound. Finally, $\|P_s-P_t\|_2$ is the largest sine of the principal angles between the two rank-$k$ subspaces; averaging their squared cosines yields~\eqref{eq:fisher-overlap}.
\end{proof}

\textbf{Corollary 1 (overlap implies target distributional sensitivity).}
Let $v$ be uniformly distributed on the unit sphere in $\mathrm{range}(P_s)$. Then
\begin{equation}\label{eq:target-fisher-energy}
    \mathbb{E}_v[v^\top F_t v]
    =\frac{1}{k}\mathrm{tr}(P_sF_t)
    \geq\lambda_k(F_t)\Omega_k.
\end{equation}
The equality follows from $\mathbb{E}_v[vv^\top]=P_s/k$, and the inequality follows from $F_t\succeq\lambda_k(F_t)P_t$.
Thus, a random unit direction in the source top-$k$ subspace has large target Fisher sensitivity on average when both $\Omega_k$ and $\lambda_k(F_t)$ are large, providing a direct quantitative link from surrogate-subspace directions to target predictive-distribution change. Proposition~4 next supplies the oriented, direction-specific condition for target-margin crossing.

\textbf{Proposition 4 (local transfer guarantee).}
Let $h_s\neq0$ be an actual feasible local displacement produced by the source attack, $d_s=h_s/\|h_s\|_2$, and let $a_{t,c}(x)=\ell_{t,c}(x)-\ell_{t,y}(x)$ be the target logit advantage of some class $c\neq y$, where prediction is by $\operatorname*{arg\,max}_j\ell_{t,j}$. Set $g_t=\nabla a_{t,c}(x)\neq0$ and $\widehat g_t=g_t/\|g_t\|_2$. Suppose that $a_{t,c}$ is differentiable and its gradient is $\beta_t$-Lipschitz on a neighborhood containing $\{x+\tau h_s:\tau\in[0,1]\}$ and that, for some $0\leq\eta_s,\eta_t<1$ and $0<\rho\leq1$,
\begin{equation}\label{eq:oriented-coherence}
\begin{aligned}
    \|(I-P_\star)d_s\|_2&\leq\eta_s,\qquad
    \|(I-P_\star)\widehat g_t\|_2\leq\eta_t,\\
    \langle P_\star d_s,P_\star\widehat g_t\rangle
    &\geq\rho\|P_\star d_s\|_2\|P_\star\widehat g_t\|_2.
\end{aligned}
\end{equation}
For example, if $d_s$ and $\widehat g_t$ have residual norms at most $\zeta_s$ and $\zeta_t$ in $\mathrm{range}(P_s)$ and $\mathrm{range}(P_t)$, respectively, then, provided that $\zeta_m+\xi_m/(g_k-\xi_m)<1$, Proposition~3 and the triangle inequality permit the choice
\begin{equation}
    \eta_m=\zeta_m+\frac{\xi_m}{g_k-\xi_m},
    \qquad m\in\{s,t\}.
\end{equation}
Define
\begin{equation}
    \kappa=\rho\sqrt{1-\eta_s^2}\sqrt{1-\eta_t^2}-\eta_s\eta_t.
\end{equation}
If $\kappa>0$, then
\begin{equation}\label{eq:target-margin-bound}
    a_{t,c}(x+h_s)\geq a_{t,c}(x)
    +\|h_s\|_2\|g_t\|_2\kappa
    -\frac{\beta_t}{2}\|h_s\|_2^2.
\end{equation}
In particular, a positive right-hand side guarantees that the target predicts a class other than $y$.

\begin{proof}
Decompose $d_s$ and $\widehat g_t$ into their mutually orthogonal $P_\star$ and $I-P_\star$ components. Equation~\eqref{eq:oriented-coherence} and Cauchy--Schwarz give $\langle d_s,\widehat g_t\rangle\geq\kappa$. The lower smoothness inequality applied to the actual displacement $h_s$ then gives~\eqref{eq:target-margin-bound}. If its right-hand side is positive, $\ell_{t,c}(x+h_s)>\ell_{t,y}(x+h_s)$.
\end{proof}

Proposition~4 converts Fisher-subspace sharing into a target-margin guarantee. Specifically, oriented within-subspace coherence aligns the actual source displacement with the target gradient, target smoothness controls finite-step curvature, and sufficient margin change yields misclassification. Together with Proposition~3, this result provides an information-geometric mechanism linking IKD's Fisher-sensitive directions to cross-model transfer by integrating the gradient-alignment and smoothness principles established in~\cite{demontis2019why,yang2021trs} into the Fisher-subspace account. By formulating the result for the actual feasible displacement $h_s$, Equations~\eqref{eq:projector-bound} through~\eqref{eq:target-margin-bound} connect the local geometry directly to finite-step momentum and projected $L_\infty$ attack outputs. \Cref{fig:proof}(b) visualizes this mechanism.

\subsubsection{Geometry Induced by MSE}
We finally compare the CE/KL-equivalent objective with MSE. For MSE,
\begin{equation}
    \mathrm{MSE}(p,q)=\|p-q\|_2^2.
\end{equation}
Using $q=p+J\delta+O(\|\delta\|_2^2)$, we obtain
\begin{equation}
    \mathrm{MSE}(p,q)=\delta^\top J^\top J\delta+O(\|\delta\|_2^3).
\end{equation}
Unlike CE and KL, MSE uses an isotropic Euclidean metric in output probability space; its pullback to input space is $J^\top J$, rather than the Fisher metric $J^\top D^{-1}J$. This pullback therefore lacks the $D^{-1}$ weighting that represents the local categorical Fisher geometry. At the gradient level, MSE produces
\begin{equation}
    \frac{\partial \mathrm{MSE}(p,q)}{\partial q_i}=2(q_i-p_i),
\end{equation}
whereas the shared CE/KL prediction-level gradient is given in~\Cref{eq:ikd-prediction-gradient}.
This difference reweights the output dimensions and interferes with the hard-label attack objective, providing a geometric explanation for the substantial performance degradation observed in the MSE ablation.

\subsubsection{Summary}
The analysis establishes three results. First, under the fixed benign prediction anchor $p=f_\varphi(x)$, soft-label CE and KL differ only by the constant entropy term $H(p)$ and induce identical gradients, Hessians, and iterative attack trajectories under matched implementations. Second, both objectives locally induce the surrogate pullback Fisher metric $J^\top D^{-1}J$. Under Assumption~1, local same-task stability relative to a Fisher eigengap yields a quantitative lower bound on dominant Fisher-subspace overlap between surrogate and target models. Corollary~1 connects this overlap to target distributional sensitivity, and Proposition~4 shows how oriented coherence and target smoothness produce a sufficient target-margin crossing condition. Third, MSE uses the Euclidean pullback $J^\top J$, providing a geometric explanation for its lower performance in our ablations. Together, these results connect IKD's output-distribution objective, shared Fisher-sensitive structure, and adversarial transfer.

\subsection{Integration with Gradient-based Attacks}
The CE/KL relationship and common input gradient are established in~\Cref{eq:ikd-ce-kl,eq:ikd-input-gradient}. Similar to previous work~\cite{xie2019improving,long2022frequency}, we integrate the IKD objective in~\Cref{equation:IKD optimal} with MIFGSM. We initialize
\begin{equation}\label{eq:mifgsm-init}
    x^{adv}_0=x,\quad g_0=0.
\end{equation}
The benign prediction anchor $p=f_\varphi(x)$ is detached and fixed during the iterations. At iteration $t$, let $q_t=f_\varphi(x^{adv}_t)$ and use the total loss in~\Cref{equation:IKD optimal}, with the KL or CE instantiation referenced in~\Cref{eq:ikd-kl,eq:ikd-ce}. The update is then
\begin{align}
    g_{t+1}&=\mu\cdot g_t+\frac{\nabla_{x^{adv}_t} \mathcal{L}_{total}}{\|\nabla_{x^{adv}_t} \mathcal{L}_{total}\|_1},\label{eq:mifgsm-momentum}\\
    x^{adv}_{t+1}&=\Pi_{\mathcal{B}_\infty(x,\epsilon)\cap\mathcal{X}}\left(x^{adv}_t+\alpha\cdot \operatorname{sign}(g_{t+1})\right).\label{eq:mifgsm-projected-step}
\end{align}
where $x^{adv}_t$ and $g_t$ denote the adversarial example and accumulated gradient at iteration $t$, respectively; $\alpha$ is the step size, $\mu$ is the momentum decay factor, and $\mathcal{X}$ is the valid input range. IKD can be combined with input transformations or gradient processing because it modifies the loss rather than those mechanisms. The detailed algorithmic workflow of MIFGSM-IKD is given in~\Cref{alg:algorithm}.

\begin{algorithm}[htbp]
    \caption{MIFGSM-IKD}
    \label{alg:algorithm}
    \textbf{Input}: Classifier $f_\varphi$; benign image $x$; label $y$; budget $\epsilon$; step size $\alpha$; iterations $T$; momentum decay $\mu$; IKD weight $\gamma$\\
    \textbf{Output}: $x_T^{adv}\in\mathcal{B}_\infty(x,\epsilon)\cap\mathcal{X}$.
    \begin{algorithmic}[1]
        \STATE $p\gets\operatorname{stopgrad}(f_\varphi(x))$; initialize $x_0^{adv},g_0$ by~\Cref{eq:mifgsm-init}.
        \FOR{$t=0$ to $T-1$}
        \STATE $q_t\gets f_\varphi(x_t^{adv})$; evaluate $\mathcal{L}_{total}$ by~\Cref{equation:IKD optimal}.
        \STATE $r_t\gets\nabla_{x_t^{adv}}\mathcal{L}_{total}$.
        \STATE Update $g_{t+1}$ with $r_t$ by~\Cref{eq:mifgsm-momentum}.
        \STATE Projected update $x_{t+1}^{adv}$ by~\Cref{eq:mifgsm-projected-step}.
        \ENDFOR
        \STATE \textbf{return} $x_T^{adv}$.
    \end{algorithmic}
\end{algorithm}

\section{Experiments}\label{Experiments}

\subsection{Experimental Setup}\label{Experiments Setting}
\textbf{Dataset.}
We follow the standard evaluation protocol of transfer-based attacks on ImageNet and adopt a widely used subset of 1,000 validation images\footnote{\sloppy\url{https://drive.google.com/drive/folders/1CfobY6i8BfqfWPHL31FKFDipNjqWwAhS}}~\cite{russakovsky2015imagenet}.  
The subset spans nearly all major ImageNet categories and is employed in many prior works, enabling a fair comparison.  
All images are resized to $3\times224\times224$ and their tensors use pixel values in $[0,1]$.  
We consider perturbations bounded in the $L_\infty$ norm with maximum budget $\epsilon=16/255$, i.e., $\|\delta\|_\infty\le16/255$ (equivalently, a budget of $16$ on the 8-bit $[0,255]$ scale).

\textbf{Evaluation models.}
We evaluate attacks on a diverse set of 15 ImageNet models, including 13 normally trained models and 2 adversarially trained defense models, all obtained from the \texttt{timm} package~\cite{rw2019timm}.  
The standard models cover both convolutional and transformer architectures and include
ResNet50 (RN50)~\cite{he2016deep},
DenseNet121 (DN121)~\cite{huang2017densely},
ResNeXt50 (RNX50)~\cite{xie2017aggregated},
VGG19BN~\cite{simonyan2015very},
InceptionResNet-v2 (IncRes-v2)~\cite{szegedy2017inception},
Inception-v3 (Inc-v3)~\cite{szegedy2016rethinking},
Inception-v4 (Inc-v4)~\cite{szegedy2017inception},
ResNet101 (RN101)~\cite{he2016identity},
ResNet152 (RN152)~\cite{he2016identity},
Visformer-S~\cite{chen2021visformer},
ViT-B/16~\cite{dosovitskiy2020vit},
Swin-B/S3~\cite{liu2021Swin},
and PiT-B~\cite{Heo_2021_ICCV}.  
For robustness evaluation, we additionally include two adversarially trained models, Inception-v3$\rm _{adv}$ and InceptionResNet-v2$\rm _{adv,ens}$~\cite{tramer2018ensemble}.

\textbf{Metrics.}
We report the attack success rate (ASR), i.e., the fraction of inputs for which the adversarial example is misclassified by the target model.  
For each source model, we report both white-box ASR (on the source itself) and black-box ASR (on all other models).  
Unless otherwise specified, the “AVG” column in tables denotes the mean ASR over all black-box targets.

\textbf{Baselines.}
To comprehensively assess IKD, we integrate it with a wide range of gradient-based transfer attacks, including
MIFGSM~\cite{dong2018boosting},
DIFGSM~\cite{xie2019improving},
TIFGSM~\cite{dong2019evading},
NIFGSM~\cite{lin2019nesterov},
SINIFGSM~\cite{lin2019nesterov},
VMIFGSM~\cite{wang2021enhancing},
VNIFGSM~\cite{wang2021enhancing},
BSR~\cite{wang2024boosting},
and GGS~\cite{niu2025enhancing}.
These cover classical momentum-based attacks, input-transformation-based strategies, and advanced gradient refinements.  
When combined with IKD, we suffix “-IKD” to indicate the corresponding variant (e.g., MIFGSM-IKD).

\textbf{Implementation details.}
Following common practice~\cite{dong2019evading,wang2021enhancing,xie2019improving}, we fix the perturbation budget to $\epsilon=16/255$ ($L_\infty$), the number of iterations to $T=10$, the step size to $\alpha=2/255$, and the momentum decay to $\mu=1.0$ where applicable.  
All attacks are implemented using the Torchattacks toolkit~\cite{kim2020torchattacks}.  
We keep all default hyperparameters of each baseline in Torchattacks, except for $\epsilon$, $\alpha$, and $T$, which are unified across methods to ensure fairness.  
Experiments are conducted with PyTorch on a single NVIDIA GeForce RTX 3090 GPU.

\subsection{Overall Transferability Results}

In this section, we first study how IKD affects transferability when combined with classical and input-transformation-based attacks, and then with advanced gradient-based attacks.  
In all tables, methods with the “-IKD” suffix use the same backbone attack and step size, but augment the loss with the KL-based IKD term described in Section~\ref{Method}.
\begin{table*}[htbp]
    \centering
    \caption{Comparison results of various attack methods in terms of attack success rate ($\%$). The bold item indicates the best one. Item with $\star$ superscript is white-box attacks, and the others are black-box attacks. AVG column indicates the average attack success rate on black-box models.}
    \resizebox{1.0\linewidth}{!}{
    {\small
    \begin{tabular}{clcccccccccccc}
    \hline
        \multirow{2}{*}{\makecell{Source \\ Model}} & \multirow{2}{*}{Method} & \multicolumn{12}{c}{Target Model} \\ \cline{3-14}
         &  & RN50 & DN121 & RNX50 & VGG19BN & IncRes-v2 & Inc-v3 & Inc-v4 & RN101 & RN152 & Inc-v3$adv$ &  IncRes-V2$adv,ens$ & AVG \\ \hline
        \multirow{8}{*}{\rotatebox{90}{RN50}} & MIFGSM & 99.9$\star$ & 64.9  & 74.1  & 73.6  & 39.1  & 52.7  & 46.5  & 68.1  & 62.9  & 31.2  & \textbf{17.1}  & 53.0  \\ 
        ~ & MIFGSM-IKD & \textbf{100.0$\star$} & \textbf{69.8}  & \textbf{77.9}  & \textbf{74.6}  & \textbf{39.9}  & \textbf{55.0}  & \textbf{48.6}  & \textbf{73.0}  & \textbf{67.7}  & \textbf{31.7}  & 16.9  & \textbf{55.5}  \\ \cline{2-14}
        ~ & DIFGSM & \textbf{99.9$\star$} & 82.9  & 89.7  & 82.2  & 59.8  & 69.9  & 66.8  & 84.1  & 81.6  & 37.7  & 18.1  & 67.3  \\
        ~ & DIFGSM-IKD & 99.6$\star$ & \textbf{85.9}  & 89.7  & \textbf{85.1}  & \textbf{64.6}  & \textbf{70.1}  & \textbf{70.9}  & \textbf{87.3}  & \textbf{84.9}  & \textbf{38.1}  & \textbf{19.0}  & \textbf{69.6}  \\ \cline{2-14}
        ~ & TIFGSM & 98.8$\star$ & 68.2  & 62.2  & 70.3  & 50.1  & 61.4  & 59.3  & 53.6  & 45.7  & 51.5  & 41.0  & 56.3  \\
        ~ & TIFGSM-IKD & \textbf{99.0$\star$} & \textbf{68.3}  & \textbf{66.9}  & \textbf{73.2}  & \textbf{52.1}  & \textbf{63.1}  & \textbf{60.8}  & \textbf{58.0}  & \textbf{52.7}  & \textbf{53.6}  & \textbf{44.2}  & \textbf{59.3}  \\ \cline{2-14}
        ~ & NIFGSM & 99.9$\star$ & 72.5  & 81.1  & 77.6  & 41.3  & 55.8  & 49.4  & 76.3  & 71.8  & 31.2  & 15.8  & 57.3  \\
        ~ & NIFGSM-IKD & 99.9$\star$ & \textbf{76.1}  & \textbf{84.0}  & \textbf{79.3}  & \textbf{44.7}  & \textbf{58.5}  & \textbf{53.3}  & \textbf{80.5}  & \textbf{76.3}  & \textbf{33.1}  & \textbf{16.1}  & \textbf{60.2}  \\ \hline
        \multirow{8}{*}{\rotatebox{90}{DN121}} & MIFGSM & 75.5  & 100.0$\star$ & 72.7  & 85.5  & 51.5  & 66.3  & 58.8  & 63.0  & 57.1  & 38.6  & \textbf{20.8}  & 59.0  \\
        ~ & MIFGSM-IKD & \textbf{79.2}  & 100.0$\star$ & \textbf{78.0}  & \textbf{88.3}  & \textbf{56.4}  & \textbf{69.8}  & \textbf{64.0}  & \textbf{67.8}  & \textbf{62.0}  & \textbf{39.9}  & 20.5  & \textbf{62.6}  \\ \cline{2-14}
        ~ & DIFGSM & 87.7  & 100.0$\star$ & 84.2  & 93.3  & 73.2  & 81.6  & 79.5  & 78.1  & 73.9  & 47.3  & 27.9  & 72.7  \\
        ~ & DIFGSM-IKD & \textbf{89.0}  & 100.0$\star$ & \textbf{86.6}  & \textbf{93.9}  & \textbf{76.8}  & \textbf{84.1}  & \textbf{83.2}  & \textbf{81.2}  & \textbf{77.0}  & \textbf{49.0}  & \textbf{30.9}  & \textbf{75.2}  \\ \cline{2-14}
        ~ & TIFGSM & 58.6  & \textbf{99.7$\star$} & \textbf{54.1}  & 80.3  & 60.2  & 70.3  & 65.5  & 40.3  & 34.4  & 64.7  & 56.4  & 58.5  \\
        ~ & TIFGSM-IKD & \textbf{60.0}  & 99.6$\star$ & 53.9  & \textbf{81.8}  & \textbf{62.2}  & \textbf{73.7}  & \textbf{70.6}  & \textbf{41.1}  & \textbf{36.7}  & \textbf{66.6}  & \textbf{57.6}  & \textbf{60.4}  \\ \cline{2-14}
        ~ & NIFGSM & 78.9  & 100.0$\star$ & 78.2  & 87.6  & 55.1  & 68.5  & 62.0  & 67.4  & 59.8  & 38.6  & 19.6  & 61.6  \\
        ~ & NIFGSM-IKD & \textbf{81.4}  & 100.0$\star$ & \textbf{80.5}  & \textbf{89.9}  & \textbf{58.8}  & \textbf{71.2}  & \textbf{65.7}  & \textbf{72.4}  & \textbf{65.5}  & \textbf{40.7}  & \textbf{20.2}  & \textbf{64.6}  \\ \hline
        \multirow{8}{*}{\rotatebox{90}{RNX50}} & MIFGSM & 68.2  & 60.8  & \textbf{98.7$\star$} & 73.2  & 32.8  & \textbf{46.8}  & 41.4  & 61.6  & 55.2  & \textbf{37.3}  & 14.7  & 49.2  \\
        ~ & MIFGSM-IKD & \textbf{71.0}  & \textbf{63.1}  & 98.6$\star$ & \textbf{73.3}  & \textbf{33.1}  & 44.9  & \textbf{43.7}  & \textbf{66.2}  & \textbf{59.5}  & 36.2  & 14.7  & \textbf{50.6}  \\ \cline{2-14}
        ~ & DIFGSM & \textbf{79.2}  & 74.2  & \textbf{96.0$\star$} & \textbf{81.4}  & 53.1  & 59.4  & 60.4  & 75.4  & 71.7  & \textbf{38.6}  & 16.2  & 61.0  \\
        ~ & DIFGSM-IKD & 78.1  & \textbf{75.1}  & 94.4$\star$ & 80.4  & \textbf{54.0}  & \textbf{61.8}  & \textbf{61.3}  & \textbf{75.6}  & \textbf{72.5}  & 37.2  & \textbf{16.8}  & \textbf{61.3}  \\ \cline{2-14}
        ~ & TIFGSM & 53.9  & 56.1  & 84.2$\star$ & 68.2  & 41.0  & 54.1  & 52.3  & 42.7  & 41.2  & 40.5  & 33.8  & 48.4  \\
        ~ & TIFGSM-IKD & \textbf{54.7}  & \textbf{56.9}  & \textbf{85.5$\star$} & \textbf{69.0}  & \textbf{42.3}  & \textbf{55.4}  & \textbf{53.0}  & \textbf{45.2}  & \textbf{42.0}  & \textbf{42.4}  & 33.8  & \textbf{49.5}  \\ \cline{2-14}
        ~ & NIFGSM & 75.3  & \textbf{69.2}  & 99.7$\star$ & 76.7  & 35.7  & 48.6  & 43.9  & 67.8  & 61.9  & 36.6  & \textbf{14.5}  & 53.0  \\
        ~ & NIFGSM-IKD & \textbf{75.7}  & 68.8  & \textbf{99.9}$\star$ & \textbf{78.0}  & \textbf{35.8}  & \textbf{48.8}  & \textbf{45.5}  & \textbf{69.2}  & \textbf{63.9}  & \textbf{37.1}  & 14.1  & \textbf{53.7}  \\ \hline
        \multirow{8}{*}{\rotatebox{90}{VGG19BN}} & MIFGSM & 47.4  & 58.4  & 43.3  & 100.0$\star$ & 30.3  & 48.9  & 44.4  & 32.5  & 30.1  & 26.7  & 14.7  & 35.7  \\
        ~ & MIFGSM-IKD & \textbf{57.6}  & \textbf{65.1}  & \textbf{51.2}  & 100.0$\star$ & \textbf{34.8}  & \textbf{54.1}  & \textbf{50.0}  & \textbf{42.1}  & \textbf{38.5}  & \textbf{28.0}  & \textbf{15.7}  & \textbf{43.7}  \\ \cline{2-14}
        ~ & DIFGSM & 62.4  & 74.7  & 56.0  & 100.0$\star$ & 50.6  & 65.1  & 62.8  & 46.9  & 45.5  & 33.0  & 17.5  & 51.5  \\
        ~ & DIFGSM-IKD & \textbf{72.0}  & \textbf{82.8}  & \textbf{69.1}  & 100.0$\star$ & \textbf{57.5}  & \textbf{71.9}  & \textbf{70.5}  & \textbf{59.8}  & \textbf{52.9}  & \textbf{35.8}  & \textbf{18.5}  & \textbf{59.1}  \\ \cline{2-14}
        ~ & TIFGSM & 39.4  & 59.7  & 35.5  & 100.0$\star$ & 41.8  & 62.0  & 56.7  & 24.4  & 21.3  & 45.8  & 33.3  & 42.0  \\
        ~ & TIFGSM-IKD & \textbf{47.7}  & \textbf{67.4}  & \textbf{40.8}  & 100.0$\star$ & \textbf{48.6}  & \textbf{68.4}  & \textbf{65.0}  & \textbf{29.0}  & \textbf{24.7}  & \textbf{50.9}  & \textbf{37.4}  & \textbf{48.0}  \\ \cline{2-14}
        ~ & NIFGSM & 55.1  & 62.4  & 46.7  & 100.0$\star$ & 33.7  & 51.4  & 45.4  & 37.4  & 35.3  & 26.0  & 14.3  & 40.8  \\
        ~ & NIFGSM-IKD & \textbf{64.1}  & \textbf{69.4}  & \textbf{58.4}  & 100.0$\star$ & \textbf{40.4}  & \textbf{54.5}  & \textbf{54.3}  & \textbf{47.3}  & \textbf{43.5}  & \textbf{28.2}  & \textbf{15.3}  & \textbf{47.5}  \\ \hline
        \multirow{8}{*}{\rotatebox{90}{IncRes-v2}} & MIFGSM & 55.5  & 67.2  & 54.9  & 77.9  & 98.7$\star$ & 73.8  & 69.9  & 48.7  & 45.3 & 44.0  & 28.6  & 56.6  \\
        ~ & MIFGSM-IKD & \textbf{60.1}  & \textbf{68.7}  & \textbf{57.9}  & \textbf{80.6}  & \textbf{99.0$\star$} & \textbf{77.9}  & \textbf{75.5}  & \textbf{51.2}  & \textbf{48.7}  & \textbf{47.3}  & \textbf{30.0}  & \textbf{59.8}  \\ \cline{2-14}
        ~ & DIFGSM & 64.4  & 75.8  & 64.5  & 83.7  & \textbf{98.7$\star$} & 83.5  & 83.6  & 58.4  & 55.2  & 54.9  & 37.3  & 66.1  \\
        ~ & DIFGSM-IKD & \textbf{70.1}  & \textbf{79.1}  & \textbf{69.1}  & \textbf{85.8}  & 97.8$\star$ & \textbf{85.8}  & \textbf{85.2}  & \textbf{62.3}  & \textbf{58.9}  & \textbf{58.2}  & \textbf{41.5}  & \textbf{69.6}  \\ \cline{2-14}
        ~ & TIFGSM & 42.7  & 58.5  & 39.8  & 67.7  & \textbf{97.8$\star$} & 74.8  & 71.3  & 31.0  & 27.1  & 68.0  & 61.9  & 54.3  \\
        ~ & TIFGSM-IKD & \textbf{44.8}  & \textbf{61.4}  & \textbf{42.2}  & \textbf{69.5}  & 97.3$\star$ & \textbf{78.0}  & \textbf{73.6}  & \textbf{33.4}  & \textbf{30.2}  & \textbf{70.2}  & \textbf{66.5}  & \textbf{57.0}  \\ \cline{2-14}
        ~ & NIFGSM & 55.8  & 66.2  & 57.0  & 78.4  & 99.6$\star$ & 72.6  & 71.0  & 50.3  & 46.1  & 43.9  & 26.5  & 56.8  \\
        ~ & NIFGSM-IKD & \textbf{59.6}  & \textbf{68.7}  & \textbf{59.5}  & \textbf{81.3}  & 99.6$\star$ & \textbf{77.8}  & \textbf{75.4}  & \textbf{51.5}  & \textbf{49.0}  & \textbf{46.4}  & \textbf{27.1}  & \textbf{59.6}  \\ \hline
        \multirow{8}{*}{\rotatebox{90}{Inc-v3}} & MIFGSM & \textbf{49.0}  & \textbf{62.3}  & \textbf{46.3}  & 72.1  & 57.7  & 99.1$\star$ & 63.0  & 39.9  & \textbf{37.0}  & 40.9  & 21.6  & \textbf{49.0}  \\
        ~ & MIFGSM-IKD & 47.5  & 60.8  & 45.4  & \textbf{72.6}  & 57.7  & \textbf{99.6$\star$} & \textbf{63.1}  & \textbf{40.5}  & 35.8  & \textbf{43.0}  & 21.6  & 48.8  \\ \cline{2-14}
        ~ & DIFGSM & 53.7  & 67.9  & 50.2  & 78.0  & 71.2  & \textbf{99.1$\star$} & 74.4  & 46.9  & 41.2  & 49.3  & 26.9  & 56.0  \\
        ~ & DIFGSM-IKD & \textbf{59.1}  & \textbf{71.2}  & \textbf{55.3}  & \textbf{80.7}  & \textbf{72.0}  & 98.9$\star$ & \textbf{77.6}  & \textbf{50.8}  & \textbf{45.2}  & \textbf{53.6}  & \textbf{28.1}  & \textbf{59.4}  \\ \cline{2-14}
        ~ & TIFGSM & 31.0  & 49.8  & 28.1  & 65.0  & 50.5  & \textbf{98.9$\star$} & 61.8  & 21.8  & 18.4  & 55.9  & 41.6  & 42.4  \\
        ~ & TIFGSM-IKD & \textbf{32.7}  & \textbf{51.2}  & \textbf{29.8}  & \textbf{66.5}  & \textbf{56.0}  & 98.7$\star$ & \textbf{63.9}  & \textbf{23.1}  & \textbf{20.7}  & \textbf{61.1}  & \textbf{46.0}  & \textbf{45.1}  \\ \cline{2-14}
        ~ & NIFGSM & 56.6  & 69.9  & 53.5  & 76.2  & 63.7  & 99.4$\star$ & 69.3  & 46.6  & 42.9  & 41.9  & 21.4  & 54.2  \\
        ~ & NIFGSM-IKD & \textbf{58.8}  & \textbf{70.4}  & \textbf{55.4}  & \textbf{78.0}  & \textbf{70.1}  & \textbf{99.7$\star$} & \textbf{73.8}  & \textbf{48.7}  & \textbf{45.1}  & \textbf{43.8}  & \textbf{22.3}  & \textbf{56.6}  \\ \hline
        \multirow{8}{*}{\rotatebox{90}{Inc-v4}} & MIFGSM & 53.2  & 61.4  & 51.9  & 76.7  & 60.9  & 69.5  & 98.0$\star$ & \textbf{46.9}  & 43.3  & \textbf{37.8}  & \textbf{19.7}  & 52.1  \\
        ~ & MIFGSM-IKD & \textbf{55.0}  & \textbf{61.6}  & \textbf{52.9}  & \textbf{79.8}  & \textbf{62.3}  & \textbf{70.7}  & \textbf{98.9$\star$} & 46.2  & \textbf{44.6}  & 37.6  & 19.6  & \textbf{53.0}  \\ \cline{2-14}
        ~ & DIFGSM & 61.6  & 71.6  & 61.9  & 81.0  & 73.3  & 77.2  & 97.9$\star$ & 54.2  & 52.0  & 44.4  & 25.5  & 60.3  \\
        ~ & DIFGSM-IKD & \textbf{65.5}  & \textbf{74.3}  & \textbf{63.5}  & \textbf{83.4}  & \textbf{77.4}  & \textbf{82.1}  & \textbf{98.8$\star$} & \textbf{57.2}  & \textbf{54.9}  & \textbf{46.5}  & \textbf{26.9}  & \textbf{63.2}  \\ \cline{2-14}
        ~ & TIFGSM & 37.6  & 50.6  & 36.4  & 67.7  & 54.8  & 65.9  & 97.3$\star$ & 27.2  & 24.4  & 53.1  & 44.1  & 46.2  \\
        ~ & TIFGSM-IKD & \textbf{39.8}  & \textbf{52.8}  & \textbf{40.0}  & \textbf{68.6}  & \textbf{58.8}  & \textbf{68.7}  & \textbf{97.6$\star$} & \textbf{29.4}  & \textbf{25.4}  & \textbf{55.1}  & \textbf{45.7}  & \textbf{48.4}  \\ \cline{2-14}
        ~ & NIFGSM & 59.6  & 67.2  & 58.3  & 81.2  & 61.9  & 72.0  & 98.6$\star$ & 50.5  & 45.9  & 38.0  & 19.6  & 55.4  \\
        ~ & NIFGSM-IKD & \textbf{61.3}  & \textbf{68.5}  & \textbf{60.7}  & \textbf{83.7}  & \textbf{65.0}  & \textbf{74.6}  & \textbf{99.7$\star$} & \textbf{53.8}  & \textbf{49.4}  & \textbf{40.0}  & \textbf{20.3}  & \textbf{57.7}  \\ \hline
        \multirow{8}{*}{\rotatebox{90}{RN101}} & MIFGSM & 72.2  & 65.2  & 71.8  & 73.9  & 36.0  & 50.1  & 44.6  & 97.0$\star$ & 70.0  & 28.4  & \textbf{15.7}  & 52.8  \\
        ~ & MIFGSM-IKD & \textbf{75.5}  & \textbf{67.3}  & \textbf{74.4}  & \textbf{75.1}  & \textbf{38.2}  & \textbf{51.7}  & \textbf{46.0}  & \textbf{97.4$\star$} & \textbf{71.3}  & \textbf{28.7}  & 15.1  & \textbf{54.3}  \\ \cline{2-14}
        ~ & DIFGSM & \textbf{78.1}  & 73.3  & 77.2  & 78.8  & 53.7  & 61.6  & \textbf{61.4}  & \textbf{92.3$\star$} & 77.5  & \textbf{33.1}  & 16.8  & 61.2  \\
        ~ & DIFGSM-IKD & 78.0  & \textbf{75.4}  & \textbf{78.5}  & \textbf{81.4}  & \textbf{56.1}  & \textbf{63.0}  & 60.8  & 91.7$\star$ & \textbf{78.5}  & 33.0  & \textbf{17.4}  & \textbf{62.2}  \\ \cline{2-14}
        ~ & TIFGSM & \textbf{52.8}  & 53.8  & \textbf{52.9}  & 64.4  & \textbf{44.3}  & 53.7  & 51.9  & 78.0$\star$ & \textbf{50.4}  & 41.3  & \textbf{35.6}  & 50.0  \\
        ~ & TIFGSM-IKD & 52.4  & \textbf{54.8}  & 52.7  & \textbf{65.3}  & 43.5  & \textbf{55.4}  & \textbf{52.1}  & \textbf{78.2$\star$} & 50.2  & \textbf{42.6}  & 35.5  & \textbf{50.5}  \\ \cline{2-14}
        ~ & NIFGSM & 79.0  & 72.7  & 77.5  & 77.3  & 38.7  & 51.4  & \textbf{46.5}  & \textbf{99.4$\star$} & 74.5  & 29.1  & 15.0  & 56.2  \\
        ~ & NIFGSM-IKD & \textbf{80.1}  & \textbf{73.9}  & \textbf{77.6}  & \textbf{78.3}  & \textbf{39.4}  & \textbf{54.2}  & 46.2  & 99.2$\star$ & \textbf{75.9}  & \textbf{29.8}  & \textbf{15.4}  & \textbf{57.1}  \\ \hline
        \multirow{8}{*}{\rotatebox{90}{RN152}} & MIFGSM & 70.9  & 62.5  & 67.0  & 73.3  & \textbf{36.6}  & 49.2  & 42.7  & 69.2  & 95.5$\star$ & 26.9  & \textbf{15.4}  & 51.4  \\
        ~ & MIFGSM-IKD & \textbf{71.2}  & \textbf{64.5}  & \textbf{68.8}  & \textbf{74.5}  & 36.0  & \textbf{49.3}  & \textbf{43.4}  & \textbf{70.3}  & \textbf{95.9$\star$} & \textbf{27.0}  & 14.8  & \textbf{52.0}  \\ \cline{2-14}
        ~ & DIFGSM & 76.4  & 71.5  & 75.9  & 78.6  & 54.2  & 62.7  & 60.5  & 77.6  & 91.7$\star$ & \textbf{34.0}  & 18.0  & 61.0  \\
        ~ & DIFGSM-IKD & \textbf{77.5}  & \textbf{72.9}  & \textbf{77.2}  & \textbf{78.7}  & \textbf{55.2}  & 62.7  & \textbf{61.9}  & \textbf{78.2}  & 91.7$\star$ & 33.8  & 18.0  & \textbf{61.6}  \\ \cline{2-14}
        ~ & TIFGSM & \textbf{51.6}  & \textbf{53.1}  & \textbf{51.3}  & \textbf{64.8}  & \textbf{44.5}  & \textbf{54.3}  & \textbf{50.9}  & 50.4  & \textbf{75.7$\star$} & \textbf{43.1}  & \textbf{36.8}  & \textbf{50.1}  \\
        ~ & TIFGSM-IKD & 51.4  & 52.4  & 50.3  & 63.3  & 42.6  & 53.3  & 50.1  & \textbf{50.7}  & 75.1$\star$ & 42.3  & 36.5  & 49.3  \\ \cline{2-14}
        ~ & NIFGSM & 77.1  & 68.4  & 73.9  & 78.4  & 38.8  & 50.2  & 43.9  & 75.7  & 99.3$\star$ & \textbf{26.7}  & 14.6  & 54.8  \\
        ~ & NIFGSM-IKD & \textbf{77.8}  & \textbf{69.8}  & \textbf{74.9}  & 78.4  & \textbf{40.0}  & \textbf{51.0}  & \textbf{45.7}  & \textbf{76.7}  & 99.3$\star$ & 26.0  & \textbf{14.9}  & \textbf{55.5} \\ \hline
    \end{tabular}
    }
    }
  \label{tab:ngmasr}%
\end{table*}

\textbf{Combination with classical attacks and attacks based on input transformations.} As shown in~\Cref{tab:ngmasr}, our method outperforms the comparison methods on the vast majority of black-box models. Specifically, we observe a significant improvement in average transferability across almost all eleven models tested. Notably, our method achieves a $6.8\%$ increase in average transferability compared to DIFGSM on the VGG19BN model, with our DIFGSM-IKD method attaining an average transferability of $57.8\%$, while DIFGSM alone achieves $51.0\%$. Furthermore, our approach also outperforms most comparison methods under defense models in terms of average attack success rate, demonstrating the effectiveness of our method in attacking defensive mechanisms. Additionally, we find that DenseNet121 is the most vulnerable model, exhibiting the highest average transferability, which challenges the conventional belief that deeper models are inherently more robust than shallower ones. This observation indicates that model robustness is more closely linked to architectural design than to depth alone. Overall, IKD delivers strong average gains while preserving the strengths of the underlying attacks.

\begin{table*}[htbp]
  \centering
  \caption{Evaluation results of the advanced attacks combined with IKD in terms of attack success rate (\%). The bold item indicates the best one. Items with $\star$ superscript are white-box attacks, and the others are black-box attacks. The AVG column indicates the average attack success rate on black-box models.}
    \resizebox{1.0\linewidth}{!}{
    {\small
    \begin{tabular}{clcccccccccccc}
        \hline
        \multirow{2}{*}{\makecell{Source \\ Model}} & \multirow{2}{*}{Method} & \multicolumn{12}{c}{Target Model} \\
        \cline{3-14}
        & & RN50  & DN121 & RNX50 & VGG19BN & IncRes-v2 & Inc-v3 & Inc-v4 & RN101 & RN152 & Inc-v3$adv$ &  IncRes-V2$adv,ens$ & AVG \\
        \hline
        \multirow{10}{*}{\rotatebox{90}{RN50}} & SINIFGSM & 99.8$\star$ & 83.0 & 84.4 & 86.1 & 53.4 & 66.6 & 60.8 & 79.4 & 75.8 & 37.3 & 17.3 & 64.4 \\
        & SINIFGSM-IKD & \textbf{99.9$\star$} & \textbf{88.0} & \textbf{91.1} & \textbf{88.0} & \textbf{58.5} & \textbf{69.8} & \textbf{66.6} & \textbf{87.7} & \textbf{84.7} & \textbf{37.1} & \textbf{18.1} & \textbf{69.0} \\
        \cline{2-14}
        & VMIFGSM & 99.7$\star$ & 85.0 & 88.5 & 85.6 & 66.1 & \textbf{74.1} & \textbf{72.9} & 86.2 & 84.6 & 43.8 & 22.2 & 70.9 \\
        & VMIFGSM-IKD & \textbf{99.8$\star$} & \textbf{86.7} & \textbf{90.6} & \textbf{87.0} & \textbf{66.5} & 74.0 & 72.3 & \textbf{87.9} & \textbf{86.0} & \textbf{44.2} & \textbf{23.8} & \textbf{71.9} \\
        \cline{2-14}
        & VNIFGSM & 99.9$\star$ & 86.4 & 90.1 & 87.4 & 68.0 & 73.8 & 73.4 & 87.1 & 85.3 & 44.0 & 22.7 & 71.8 \\
        & VNIFGSM-IKD & 99.9$\star$ & \textbf{87.3} & \textbf{91.8} & \textbf{87.8} & \textbf{68.5} & \textbf{75.4} & \textbf{75.2} & \textbf{89.3} & \textbf{87.8} & \textbf{44.6} & \textbf{23.0} & \textbf{73.1} \\
        \cline{2-14}
        & BSR & \textbf{99.4$\star$} & 89.9 & 88.8 & 92.3 & 65.7 & 76.5 & 72.5 & 81.1 & 78.5 & 43.0 & 21.2 & 71.0 \\
        & BSR-IKD & 98.2$\star$ & \textbf{92.6} & \textbf{94.7} & \textbf{92.7} & \textbf{73.3} & \textbf{79.7} & \textbf{80.0} & \textbf{90.1} & \textbf{87.9} & \textbf{44.3} & \textbf{23.4} & \textbf{75.9} \\
        \hline
        \multirow{6}{*}{\rotatebox{90}{DN121}} & SINIFGSM & 82.4 & 100.0$\star$ & 78.6 & 92.8 & 61.4 & 76.6 & 70.5 & 69.8 & 63.3 & 47.7 & 28.0 & 67.1 \\
        & SINIFGSM-IKD & \textbf{86.9} & 100.0$\star$ & \textbf{83.5} & \textbf{94.2} & \textbf{68.2} & \textbf{80.2} & \textbf{77.1} & \textbf{76.6} & \textbf{72.3} & \textbf{49.4} & \textbf{28.2} & \textbf{71.7} \\
        \cline{2-14}
        & VMIFGSM & 87.6 & 100.0$\star$ & 85.6 & 93.0 & 73.6 & 81.5 & 80.4 & 78.8 & 76.2 & 48.0 & 33.5 & 73.8 \\
        & VMIFGSM-IKD & \textbf{88.9} & 100.0$\star$ & \textbf{88.4} & \textbf{94.2} & \textbf{74.1} & \textbf{82.5} & \textbf{81.1} & \textbf{82.0} & \textbf{78.3} & \textbf{49.2} & \textbf{35.1} & \textbf{75.4} \\
        \cline{2-14}
        & VNIFGSM & 88.2 & 100.0$\star$ & 87.3 & \textbf{94.0} & 73.9 & 82.8 & 81.2 & 81.4 & 78.9 & 49.6 & 34.3 & 75.2 \\
        & VNIFGSM-IKD & \textbf{89.3} & 100.0$\star$ & \textbf{87.9} & 93.8 & \textbf{75.6} & \textbf{83.3} & \textbf{81.6} & \textbf{81.7} & \textbf{80.5} & \textbf{50.7} & \textbf{35.1} & \textbf{76.0} \\
        \cline{2-14}
        & BSR & 89.3 & 99.9$\star$ & 85.0 & 97.3 & 74.6 & 86.2 & 82.8 & 76.4 & 72.8 & 52.0 & 31.3 & 74.8 \\
        & BSR-IKD & \textbf{93.0} & \textbf{100.0$\star$} & \textbf{90.2} & \textbf{98.4} & \textbf{78.1} & \textbf{88.9} & \textbf{86.0} & \textbf{83.7} & \textbf{78.9} & \textbf{56.3} & \textbf{33.8} & \textbf{78.7} \\
        \hline
        \multirow{6}{*}{\rotatebox{90}{RNX50}} & SINIFGSM & \textbf{91.3} & 89.0 & 100.0$\star$ & 87.4 & \textbf{61.1} & 69.3 & \textbf{68.6} & 87.6 & 83.9 & 38.5 & 17.9 & \textbf{69.5} \\
        & SINIFGSM-IKD & 90.9 & \textbf{89.4} & 100.0$\star$ & 87.4 & 60.3 & \textbf{70.4} & 66.6 & \textbf{87.7} & \textbf{84.5} & \textbf{38.9} & \textbf{18.3} & 69.4 \\
        \cline{2-14}
        & VMIFGSM & 81.7 & \textbf{79.4} & \textbf{97.8$\star$} & 80.6 & \textbf{59.2} & 67.6 & \textbf{67.8} & \textbf{79.7} & 76.5 & \textbf{41.1} & \textbf{23.3} & \textbf{65.7} \\
        & VMIFGSM-IKD & \textbf{82.6} & 77.9 & 97.7$\star$ & \textbf{80.9} & 59.1 & \textbf{67.7} & 66.5 & 78.3 & \textbf{76.7} & 40.6 & 22.6 & 65.3 \\
        \cline{2-14}
        & VNIFGSM & 81.7 & \textbf{79.1} & 96.2$\star$ & 81.4 & 60.7 & 68.1 & \textbf{68.2} & \textbf{79.2} & 77.2 & \textbf{41.3} & 23.8 & 66.1 \\
        & VNIFGSM-IKD & \textbf{82.8} & 79.0 & \textbf{98.5$\star$} & 81.4 & \textbf{61.2} & \textbf{69.8} & 67.1 & 79.1 & \textbf{78.2} & 41.0 & \textbf{24.0} & \textbf{66.4} \\
        \cline{2-14}
        & BSR & 90.1 & 89.2 & \textbf{97.5$\star$} & 89.8 & 65.3 & 75.2 & 73.9 & 82.9 & 81.9 & \textbf{42.2} & 21.1 & 71.2 \\
        & BSR-IKD & \textbf{90.9} & 89.2 & 97.4$\star$ & \textbf{90.2} & \textbf{67.1} & \textbf{75.5} & \textbf{76.9} & \textbf{85.0} & \textbf{85.1} & 40.1 & \textbf{23.2} & \textbf{72.3} \\
        \hline
        \multirow{6}{*}{\rotatebox{90}{VGG19BN}} & SINIFGSM & 55.0 & 67.4 & 48.8 & 100.0$\star$ & 38.4 & 57.6 & 49.6 & 38.5 & 34.8 & 31.1 & \textbf{16.3} & 43.8 \\
        & SINIFGSM-IKD & \textbf{69.0} & \textbf{76.2} & \textbf{61.7} & 100.0$\star$ & \textbf{45.6} & \textbf{63.0} & \textbf{58.7} & \textbf{53.2} & \textbf{47.3} & \textbf{32.2} & 15.3 & \textbf{52.2} \\
        \cline{2-14}
        & VMIFGSM & 69.2 & 81.4 & 62.1 & 100.0$\star$ & 51.4 & 70.4 & 66.2 & 53.8 & 48.9 & 37.3 & 19.9 & 56.1 \\
        & VMIFGSM-IKD & \textbf{74.2} & \textbf{83.3} & \textbf{67.9} & 100.0$\star$ & \textbf{52.7} & \textbf{71.8} & \textbf{70.0} & \textbf{59.9} & \textbf{54.0} & \textbf{36.7} & \textbf{20.9} & \textbf{59.1} \\
        \cline{2-14}
        & VNIFGSM & 71.0 & 82.6 & 64.1 & 100.0$\star$ & 53.5 & 71.1 & 69.2 & 54.4 & 48.4 & 37.3 & \textbf{20.1} & 57.2 \\
        & VNIFGSM-IKD & \textbf{76.8} & \textbf{85.7} & \textbf{72.0} & 100.0$\star$ & \textbf{57.8} & \textbf{73.7} & \textbf{71.7} & \textbf{62.2} & \textbf{58.1} & \textbf{38.2} & 19.8 & \textbf{61.6} \\
        \cline{2-14}
        & BSR & 60.6 & 72.3 & 54.2 & 99.9$\star$ & 44.0 & 63.5 & 59.3 & 45.1 & 41.4 & \textbf{37.6} & 17.5 & 49.6 \\
        & BSR-IKD & \textbf{75.5} & \textbf{81.2} & \textbf{70.4} & \textbf{100.0$\star$} & \textbf{54.7} & \textbf{70.8} & \textbf{72.5} & \textbf{62.4} & \textbf{60.0} & 34.7 & \textbf{18.6} & \textbf{60.1} \\
        \hline
        \multirow{6}{*}{\rotatebox{90}{IncRes-v2}} & SINIFGSM & 64.1 & 76.7 & 62.6 & 86.5 & 99.7$\star$ & 87.2 & 80.7 & 55.7 & 51.2 & 61.5 & 39.9 & 66.6 \\
        & SINIFGSM-IKD & \textbf{72.0} & \textbf{82.9} & \textbf{71.2} & \textbf{89.6} & \textbf{99.9$\star$} & \textbf{91.0} & \textbf{87.1} & \textbf{64.0} & \textbf{61.3} & \textbf{65.2} & \textbf{44.9} & \textbf{73.0} \\
        \cline{2-14}
        & VMIFGSM & 64.4 & 71.9 & 64.0 & 82.3 & 98.9$\star$ & 82.5 & 82.8 & 56.4 & 55.0 & 56.5 & 42.7 & 65.9 \\
        & VMIFGSM-IKD & \textbf{70.0} & \textbf{76.7} & \textbf{69.1} & \textbf{84.7} & 98.9$\star$ & \textbf{85.9} & \textbf{85.9} & \textbf{62.4} & \textbf{61.4} & \textbf{61.9} & \textbf{45.9} & \textbf{70.4} \\
        \cline{2-14}
        & VNIFGSM & 65.3 & 75.0 & 66.3 & 83.2 & 99.5$\star$ & 85.0 & 85.3 & 59.7 & 58.6 & 58.2 & 42.6 & 67.9 \\
        & VNIFGSM-IKD & \textbf{73.0} & \textbf{78.9} & \textbf{71.2} & \textbf{86.1} & 99.5$\star$ & \textbf{87.3} & \textbf{87.2} & \textbf{66.8} & \textbf{64.1} & \textbf{63.0} & \textbf{46.6} & \textbf{72.4} \\
        \cline{2-14}
        & BSR & 62.1 & 77.3 & 59.6 & 89.1 & 92.2$\star$ & 80.6 & 74.0 & 52.4 & 46.7 & 49.8 & 27.7 & 61.9 \\
        & BSR-IKD & \textbf{73.9} & \textbf{85.0} & \textbf{70.1} & \textbf{91.4} & \textbf{97.5$\star$} & \textbf{87.2} & \textbf{84.9} & \textbf{62.1} & \textbf{57.6} & \textbf{56.2} & \textbf{33.8} & \textbf{70.2} \\
        \hline
        \multirow{6}{*}{\rotatebox{90}{Inc-v3}} & SINIFGSM & 56.4 & 72.0 & 54.4 & 82.8 & 67.0 & 100.0$\star$ & 73.6 & 45.8 & 43.7 & 49.4 & 25.3 & 57.0 \\
        & SINIFGSM-IKD & \textbf{61.1} & \textbf{74.9} & \textbf{59.4} & \textbf{84.9} & \textbf{74.0} & 100.0$\star$ & \textbf{77.6} & \textbf{52.3} & \textbf{47.3} & \textbf{55.2} & \textbf{26.2} & \textbf{61.3} \\
        \cline{2-14}
        & VMIFGSM & 57.0 & 68.3 & 53.7 & 77.8 & 72.0 & 99.4$\star$ & 75.9 & 47.2 & 44.4 & 53.8 & 28.8 & 57.9 \\
        & VMIFGSM-IKD & \textbf{60.1} & \textbf{70.9} & \textbf{56.9} & \textbf{80.1} & \textbf{75.7} & \textbf{99.5$\star$} & \textbf{78.7} & \textbf{51.1} & \textbf{47.4} & \textbf{56.9} & \textbf{31.9} & \textbf{61.0} \\
        \cline{2-14}
        & VNIFGSM & 58.6 & 72.1 & 57.3 & 81.2 & 74.8 & 99.6$\star$ & 78.4 & 49.6 & 47.3 & 54.4 & 29.4 & 60.3 \\
        & VNIFGSM-IKD & \textbf{64.8} & \textbf{75.3} & \textbf{62.2} & \textbf{83.1} & \textbf{79.7} & \textbf{99.9$\star$} & \textbf{83.1} & \textbf{56.1} & \textbf{52.9} & \textbf{58.8} & \textbf{32.8} & \textbf{64.8} \\
        \cline{2-14}
        & BSR & 55.9 & 73.9 & 52.6 & 86.3 & 63.4 & 95.0$\star$ & 72.5 & 43.9 & 41.3 & 47.6 & 24.2 & 56.2 \\
        & BSR-IKD & \textbf{59.9} & \textbf{77.2} & \textbf{56.7} & \textbf{86.9} & \textbf{68.8} & \textbf{97.7$\star$} & \textbf{77.8} & \textbf{47.9} & \textbf{46.4} & \textbf{49.2} & \textbf{25.9} & \textbf{59.7} \\
        \hline
        \multirow{6}{*}{\rotatebox{90}{Inc-v4}} & SINIFGSM & 70.6 & 80.6 & 68.5 & 90.7 & 80.9 & 88.4 & 100.0$\star$ & 62.1 & 59.8 & 59.1 & 34.8 & 69.6 \\
        & SINIFGSM-IKD & \textbf{74.5} & \textbf{83.5} & \textbf{71.2} & \textbf{91.9} & \textbf{85.9} & \textbf{90.7} & 100.0$\star$ & \textbf{66.2} & \textbf{62.5} & \textbf{60.3} & \textbf{36.2} & \textbf{72.3} \\
        \cline{2-14}
        & VMIFGSM & 61.4 & 69.8 & 60.7 & 80.1 & 75.3 & 78.4 & 98.5$\star$ & 56.5 & 54.2 & 50.6 & 30.3 & 61.7 \\
        & VMIFGSM-IKD & \textbf{65.0} & \textbf{73.6} & \textbf{65.5} & \textbf{84.0} & \textbf{78.9} & \textbf{83.1} & \textbf{99.0$\star$} & \textbf{60.5} & \textbf{58.0} & \textbf{52.9} & \textbf{31.2} & \textbf{65.3} \\
        \cline{2-14}
        & VNIFGSM & 65.3 & 73.5 & 64.5 & 84.9 & 77.8 & 82.9 & 99.1$\star$ & 59.7 & 54.6 & 50.9 & 30.1 & 64.1 \\
        & VNIFGSM-IKD & \textbf{68.6} & \textbf{78.0} & \textbf{70.3} & \textbf{86.8} & \textbf{81.9} & \textbf{85.4} & \textbf{99.5$\star$} & \textbf{64.1} & \textbf{61.2} & \textbf{54.3} & \textbf{32.1} & \textbf{68.3} \\
        \cline{2-14}
        & BSR & 59.5 & 76.1 & 57.1 & 89.2 & 63.4 & 76.3 & 94.1$\star$ & 49.7 & 43.7 & 43.5 & 23.4 & 58.2 \\
        & BSR-IKD & \textbf{66.2} & \textbf{77.5} & \textbf{63.8} & \textbf{91.8} & \textbf{68.6} & \textbf{79.8} & \textbf{98.3$\star$} & \textbf{55.6} & \textbf{49.9} & \textbf{45.3} & \textbf{26.5} & \textbf{62.5} \\
        \hline
        \multirow{6}{*}{\rotatebox{90}{RN101}} & SINIFGSM & 90.3 & \textbf{87.7} & 89.6 & 87.5 & 63.2 & \textbf{72.1} & \textbf{72.3} & 99.8$\star$ & 89.9 & 37.1 & 18.8 & 70.9 \\
        & SINIFGSM-IKD & \textbf{91.3} & 87.6 & \textbf{90.9} & \textbf{88.5} & \textbf{63.8} & 71.4 & 71.3 & 99.8$\star$ & \textbf{90.4} & \textbf{37.5} & \textbf{19.6} & \textbf{71.2} \\
        \cline{2-14}
        & VMIFGSM & 79.0 & 74.2 & 78.2 & \textbf{79.8} & 58.1 & \textbf{66.0} & \textbf{63.8} & 95.5$\star$ & 78.4 & 38.4 & \textbf{25.0} & 64.1 \\
        & VMIFGSM-IKD & \textbf{80.1} & \textbf{76.6} & \textbf{78.8} & 79.7 & \textbf{59.5} & 65.1 & 63.5 & \textbf{95.8$\star$} & 78.4 & \textbf{39.4} & 23.5 & \textbf{64.5} \\
        \cline{2-14}
        & VNIFGSM & 80.4 & 77.0 & 79.6 & 80.7 & 60.4 & 67.3 & 65.2 & 95.2$\star$ & \textbf{81.4} & 38.7 & 23.4 & 65.4 \\
        & VNIFGSM-IKD & \textbf{80.6} & 77.0 & \textbf{81.8} & \textbf{81.9} & \textbf{61.2} & \textbf{67.5} & \textbf{66.8} & \textbf{96.6$\star$} & 81.1 & 38.7 & \textbf{24.1} & \textbf{66.1} \\
        \cline{2-14}
        & BSR & 91.2 & 88.8 & 90.2 & \textbf{90.8} & 69.8 & 77.0 & 77.3 & \textbf{95.3$\star$} & 87.3 & 42.9 & 22.3 & 73.8 \\
        & BSR-IKD & \textbf{91.9} & \textbf{90.0} & \textbf{91.5} & 90.2 & \textbf{73.4} & 77.0 & \textbf{79.7} & 95.1$\star$ & \textbf{88.5} & \textbf{43.1} & \textbf{23.3} & \textbf{74.9} \\
        \hline
        \multirow{6}{*}{\rotatebox{90}{RN152}} & SINIFGSM & \textbf{91.2} & 86.6 & 88.8 & 87.5 & \textbf{64.0} & \textbf{71.2} & \textbf{70.0} & \textbf{90.4} & \textbf{99.6$\star$} & \textbf{41.1} & 20.0 & \textbf{71.1} \\
        & SINIFGSM-IKD & 90.9 & \textbf{87.2} & \textbf{89.3} & \textbf{87.6} & 63.7 & 71.0 & 69.5 & 90.2 & 99.5$\star$ & 39.2 & 20.0 & 70.9 \\
        \cline{2-14}
        & VMIFGSM & 78.5 & 71.8 & \textbf{78.2} & 79.1 & 56.3 & 64.6 & \textbf{61.0} & 78.0 & 96.3$\star$ & 38.4 & \textbf{24.1} & 63.0 \\
        & VMIFGSM-IKD & \textbf{78.8} & \textbf{72.5} & 77.0 & 79.1 & \textbf{58.4} & \textbf{65.4} & 60.5 & \textbf{79.2} & \textbf{96.4$\star$} & \textbf{39.0} & 23.6 & \textbf{63.4} \\
        \cline{2-14}
        & VNIFGSM & 80.4 & 76.7 & 80.9 & \textbf{82.2} & 59.5 & \textbf{67.7} & 64.3 & 80.3 & 96.6$\star$ & \textbf{40.9} & \textbf{23.8} & 65.7 \\
        & VNIFGSM-IKD & \textbf{81.9} & \textbf{77.1} & \textbf{81.7} & 80.5 & \textbf{60.1} & 67.0 & \textbf{65.2} & \textbf{81.0} & \textbf{97.6$\star$} & 40.7 & 23.7 & \textbf{65.9} \\
        \cline{2-14}
        & BSR & 90.9 & 87.0 & 89.3 & \textbf{89.7} & 68.6 & 76.6 & 77.6 & 88.1 & 94.7$\star$ & \textbf{43.2} & 23.1 & 73.4 \\
        & BSR-IKD & \textbf{91.9} & \textbf{87.7} & \textbf{90.2} & 89.4 & \textbf{70.8} & 76.6 & \textbf{78.1} & \textbf{89.9} & 94.7$\star$ & 42.5 & \textbf{23.7} & \textbf{74.1} \\
        \hline
    \end{tabular}%
    }
    }
  \label{tab:agmasr}%
\end{table*}%

\textbf{Combination with advanced gradient attacks.} \Cref{tab:agmasr} shows that integrating IKD with advanced gradient attacks increases the black-box AVG in 33 of 36 combinations of attacks and sources. For BSR, the largest average gain is $10.5$ percentage points with VGG19BN as the source, and the largest target-specific gain is $18.6$ points on RN152. IKD also strengthens already competitive baselines such as VMIFGSM and VNIFGSM. These results demonstrate that IKD complements advanced gradient optimization and broadly strengthens black-box transfer.

\begin{table}[htbp]
    \centering
    \caption{Attack success rates (\%) of different methods with and without IKD on four ViT models. The source model is ResNet50.}
    \label{tab:c2v_results}
    \begin{tabular}{lcccc}
        \hline
        \multirow{2}{*}{Method} & \multicolumn{4}{c}{Target Model} \\
        \cline{2-5}
         & Visformer-S & ViT-B/16 & Swin-B/S3 & PiT-B \\
        \hline
        MIFGSM         & 40.6 & 19.7 & 26.3 & 34.3 \\
        MIFGSM-IKD     & \textbf{44.3} & \textbf{20.9} & \textbf{30.7} & \textbf{38.4} \\
        \hline
        NIFGSM         & 42.5 & 18.6 & 29.6 & 38.7 \\
        NIFGSM-IKD     & \textbf{46.5} & \textbf{21.1} & \textbf{31.6} & \textbf{42.3} \\
        \hline
        SINIFGSM       & 48.9 & 22.5 & 27.0 & 38.0 \\
        SINIFGSM-IKD   & \textbf{58.8} & \textbf{28.0} & \textbf{38.5} & \textbf{48.2} \\
        \hline
        VMIFGSM        & 68.7 & 38.9 & 49.8 & 56.7 \\
        VMIFGSM-IKD    & \textbf{70.1} & \textbf{39.8} & \textbf{50.9} & \textbf{60.6} \\
        \hline
        VNIFGSM        & 69.7 & 40.4 & 50.9 & 58.9 \\
        VNIFGSM-IKD    & \textbf{71.8} & \textbf{41.5} & \textbf{53.0} & \textbf{61.9} \\
        \hline
        BSR            & 71.6 & 41.8 & 47.2 & 60.2 \\
        BSR-IKD        & \textbf{81.7} & \textbf{50.5} & \textbf{58.0} & \textbf{70.3} \\
        \hline
        GGS            & 81.8 & 44.9 & 55.4 & 64.6 \\
        GGS-IKD        & \textbf{93.9} & \textbf{64.7} & \textbf{76.9} & \textbf{83.9} \\
        \hline
        
    \end{tabular}
\end{table}

\begin{table*}[htbp]
    \centering
    \caption{Attack success rates (\%) of different methods with and without IKD on CNN and ViT family models. The source model is MobileNet.}
    \label{tab:small_results}
    \resizebox{\textwidth}{!}{
    {\Huge
    \begin{tabular}{lccccccccccccc}
        \hline
        \multirow{2}{*}{Method} & \multicolumn{13}{c}{Target Model} \\
        \cline{2-14}
         & RN50 & DN121 & RNX50 & VGG19BN & IncRes-v2 & Inc-v3 & Inc-v4 & RN101 & RN152 & Visformer-S & ViT-B/16 & Swin-B/S3 & PiT-B \\
         \hline
        MIFGSM        & 35.3 & 57.1 & 30.3 & 78.9 & 30.4 & 52.2 & 41.1 & 22.8 & 20.1 & 25.0 & 13.1 & 16.2 & 17.9 \\
        MIFGSM-IKD    & \textbf{38.1} & \textbf{60.6} & \textbf{32.2} & \textbf{81.7} & \textbf{35.2} & \textbf{55.6} & \textbf{43.2} & \textbf{25.9} & \textbf{22.3} & \textbf{29.0} & \textbf{13.3} & \textbf{18.2} & \textbf{20.6} \\
        \hline
        DIFGSM        & 52.7 & 76.3 & 45.1 & 89.8 & 50.2 & 69.7 & 59.7 & 37.6 & 32.3 & 37.4 & 20.7 & 23.5 & 25.6 \\
        DIFGSM-IKD    & \textbf{56.9} & \textbf{79.0} & \textbf{49.9} & \textbf{90.8} & \textbf{54.0} & \textbf{72.2} & \textbf{64.4} & \textbf{42.2} & \textbf{35.3} & \textbf{42.6} & \textbf{22.1} & \textbf{25.4} & \textbf{29.0} \\
        \hline
        TIFGSM        & 32.8 & 59.5 & 30.0 & 76.2 & 41.5 & 63.0 & 52.2 & 18.7 & 15.1 & \textbf{21.7} & 12.3 & \textbf{6.6} & 10.9 \\
        TIFGSM-IKD    & \textbf{35.1} & \textbf{61.9} & 30.0 & \textbf{78.8} & \textbf{43.1} & \textbf{66.4} & \textbf{54.4} & \textbf{20.4} & \textbf{17.1} & 21.6 & \textbf{12.4} & 5.5 & \textbf{11.9} \\
        \hline
        NIFGSM        & 38.6 & 61.3 & 33.5 & 82.2 & 35.8 & 53.8 & 43.1 & 26.0 & 21.9 & 25.6 & 14.2 & \textbf{19.8} & 20.3 \\
        NIFGSM-IKD    & \textbf{44.5} & \textbf{66.0} & \textbf{36.6} & \textbf{87.0} & \textbf{37.2} & \textbf{58.3} & \textbf{45.1} & \textbf{29.0} & \textbf{23.2} & \textbf{29.7} & \textbf{15.2} & 19.0 & \textbf{21.9} \\
        \hline
    \end{tabular}
    }
    }
\end{table*}

\begin{table*}[htbp]
    \centering
    \caption{Attack success rates (\%) of SINIFGSM and SINIFGSM-IKD under different source models (Visformer-S and ViT-B/16) against CNN and ViT target models. Best results (IKD-enhanced) are highlighted in bold.}
    \label{tab:v2c_results}
    \resizebox{\textwidth}{!}{
    {\Huge
    \begin{tabular}{lccccccccccccc}
        \hline
        \multirow{2}{*}{\makecell{Source \\ Model}} & \multicolumn{13}{c}{Target Model} \\
        \cline{2-14}
         & RN50 & DN121 & RNX50 & VGG19BN & IncRes-v2 & Inc-v3 & Inc-v4 & RN101 & RN152 & Visformer-S & ViT-B/16 & Swin-B/S3 & PiT-B \\
        \hline
        \multirow{2}{*}{Visformer-S} 
        & 85.0 & 87.6 & 85.2 & 93.3 & 73.6 & \textbf{84.0} & 81.7 & 79.8 & 76.3 & 100.0$\star$ & 61.7 & 79.5 & 82.9 \\
        & \textbf{88.4} & \textbf{89.7} & \textbf{86.7} & \textbf{93.6} & \textbf{75.1} & 83.6 & \textbf{84.4} & \textbf{82.6} & \textbf{80.3} & 100.0$\star$ & \textbf{64.0} & \textbf{81.4} & \textbf{84.6} \\
        \hline
        \multirow{2}{*}{ViT-B/16} 
        & 54.8 & 66.1 & 53.1 & 79.2 & 53.6 & 64.1 & 58.6 & 46.2 & 47.8 & 60.3 & 99.9$\star$ & 60.9 & 55.6 \\
        & \textbf{58.6} & \textbf{68.5} & \textbf{58.3} & \textbf{81.1} & \textbf{58.1} & \textbf{66.1} & \textbf{62.0} & \textbf{51.7} & \textbf{53.0} & \textbf{63.9} & \textbf{100.0$\star$} & \textbf{64.6} & \textbf{59.1} \\
        \hline
    \end{tabular}
    }
    }
\end{table*}

\begin{figure*}[htbp]
	\centering
	\subfloat[Classical attacks and input transformation-based attacks.]{\includegraphics[width=1\columnwidth]{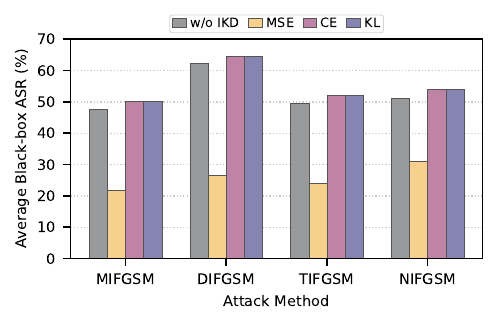}}
    \subfloat[Advanced gradient attacks.]{\includegraphics[width=1\columnwidth]{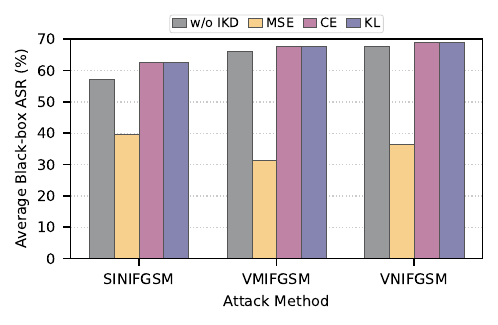}}
	\caption{Effect of different soft-label discrepancies in IKD on the average black-box attack success rate. Under the matched fixed-anchor configurations, CE and KL produce identical results. MSE significantly degrades transferability, demonstrating the disadvantage of Euclidean output matching for IKD.}
    \label{fig:regularization}
\end{figure*}

The results demonstrate that augmenting the baseline objective with IKD's prediction-distribution divergence produces more transferable adversarial examples. By enriching the surrogate-local gradient signal with additional Fisher-sensitive directions, IKD reduces surrogate overfitting and improves black-box transfer. The analysis in Section~\ref{sec:cross-model-sharing} formalizes this cross-model mechanism through dominant Fisher-subspace overlap and target-margin progress.

\subsection{Cross-Architecture Transferability}\label{sec:cross_arch}

We investigate cross-architecture transfer, where the surrogate and target belong to different model families.  
We consider both CNN $\rightarrow$ ViT and ViT $\rightarrow$ CNN settings, as well as the use of lightweight surrogates.

\textbf{CNN surrogates to ViT targets.}
Table~\ref{tab:c2v_results} reports the results when ResNet50 is used as the source model and four Vision Transformer variants (Visformer-S, ViT-B/16, Swin-B/S3, PiT-B) are targeted.  
Across all seven baseline attacks, adding IKD improves transferability to each of the four ViT targets.  
The gain is particularly striking for SINIFGSM, for which IKD raises ASR on Visformer-S from $48.9\%$ to $58.8\%$ and on Swin-B/S3 from $27.0\%$ to $38.5\%$.  
At the high-performance end, GGS-IKD reaches $93.9\%$ ASR on Visformer-S and $83.9\%$ on PiT-B.

We further evaluate the setting where the surrogate is a compact CNN (MobileNet).  
As shown in Table~\ref{tab:small_results}, combining IKD with MobileNet-based attacks produces broad improvements across CNN and ViT targets.  
DIFGSM-IKD improves over DIFGSM by $4.2$ percentage points on ResNet50 and $5.2$ points on Visformer-S, while MIFGSM-IKD yields gains of up to $4.8$ points on CNNs and $4.0$ points on ViTs.  
These results demonstrate that IKD strengthens the transferability of adversarial examples generated by compact surrogates across architecture families.

\textbf{ViT surrogates to CNN and ViT targets.}
To assess the effectiveness of IKD for transformer-based surrogates, we consider Visformer-S and ViT-B/16 as source models and evaluate transferability to both CNN and ViT targets.  
Table~\ref{tab:v2c_results} presents the results for SINIFGSM and SINIFGSM-IKD.  
IKD improves 11 of the 12 black-box targets with Visformer-S as the surrogate and all 12 black-box targets with ViT-B/16 as the surrogate.

With Visformer-S as the surrogate, the gains reach $4.0$ percentage points; with ViT-B/16, IKD improves ResNet50 and DenseNet121 by $3.8$ and $2.4$ points, respectively.  
These results demonstrate that IKD strengthens transfer from both CNN and ViT surrogates and across both within-family and cross-family targets.  
The bidirectional gains establish the broad architectural compatibility of IKD.

Overall, the cross-architecture experiments demonstrate that IKD improves transfer in both CNN$\rightarrow$ViT and ViT$\rightarrow$CNN directions and remains effective with compact surrogates. Together with the analysis in Section~\ref{sec:cross-model-sharing}, these results support the information-geometric account that enriching the surrogate gradient signal with Fisher-sensitive directions promotes transfer across heterogeneous models.

\begin{figure*}[htbp]
	\centering
	\subfloat[Classical attacks and input transformation-based attacks.]{\includegraphics[width=1\columnwidth]{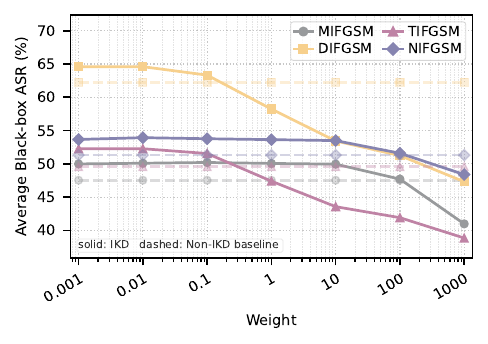}}
	\subfloat[Advanced gradient attacks.]{\includegraphics[width=1\columnwidth]{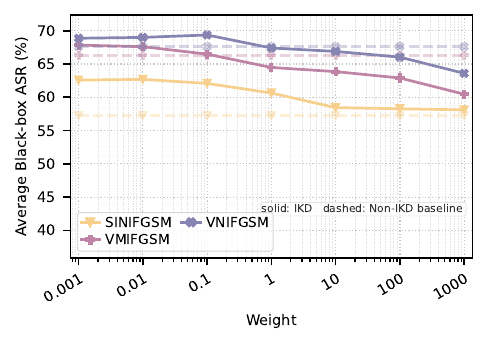}}
	\caption{Effect of IKD weight on the transfer-based average attack success rate.}
    \label{fig:weight}
\end{figure*}

\subsection{Ablation Study}\label{Ablation Study}
To identify the factors governing IKD performance, we conduct two ablation experiments. They examine (1) the effect of the soft-label discrepancy used in inverse knowledge distillation and (2) the impact of the IKD weight.

\textbf{Influence of soft-label discrepancy on IKD.} 
To investigate the effect of different soft-label discrepancies on adversarial transferability, we compare Mean Squared Error (MSE), soft-label cross-entropy (CE), and KL divergence under matched fixed-anchor configurations. Specifically, we conduct experiments on ResNet50 (RN50) using MSE, CE, and KL as $\mathcal{L}_{soft}$ in IKD. The evaluation results are presented in~\Cref{fig:regularization}. CE and KL produce exactly the same attack success rates across all seven attacks and all eleven reported model columns, in agreement with the optimization equivalence established in Proposition~1. In contrast, MSE causes substantial performance degradation, including large white-box drops for several attacks. This pronounced gap identifies output-space geometry as a decisive design factor for IKD. CE and KL induce the local Fisher pullback $J^\top D^{-1}J$, whereas MSE induces the local Euclidean pullback $J^\top J$.

\textbf{Influence of the IKD weight.}
The CE/KL-equivalent IKD objective exhibits varying influences on the gradient direction depending on the weight parameter $\gamma$, which in turn affects adversarial transferability. To investigate this relationship, we perform a grid search over $\gamma$ in the range $\{0.001, 0.01, 0.1, 1, 10, 100, 1000\}$, using ResNet50 (RN50) as the surrogate model. The average evaluation results are presented in~\Cref{fig:weight}. The sweep reveals a stable small-weight regime and a clear decline at large values of $\gamma$, where the soft term receives substantially greater weight relative to the hard-label objective. Based on these results, we select $\gamma=0.01$ as the default IKD weight.


\section{Conclusion and Future Work}\label{Conclusion}
We present Inverse Knowledge Distillation (IKD), a plug-in objective for improving the transferability of gradient-based adversarial examples. By maximizing the discrepancy between benign and adversarial output distributions, IKD enriches the surrogate-local gradient signal beyond the hard-label objective and improves transferability. We prove that soft-label cross-entropy and KL divergence differ only by a constant entropy term under the matched fixed-anchor formulation and therefore yield identical optimization trajectories. Our information-geometric analysis further shows that local same-task stability relative to a Fisher eigengap yields a quantitative lower bound on dominant Fisher-subspace overlap between surrogate and target models, connects this overlap to target distributional sensitivity, and derives a sufficient target-margin crossing condition under oriented coherence and target smoothness. These results establish a theoretical link between the Fisher-sensitive directions promoted by IKD and adversarial transfer. Extensive ImageNet experiments demonstrate substantial black-box gains across CNN, transformer, and adversarially trained targets, including bidirectional transfer between CNNs and ViTs and transfer from compact surrogates. Ablations confirm CE and KL equivalence, reveal the performance disadvantage of MSE geometry, and demonstrate a broad effective weight range in the RN50 sweep. Future work focuses on extending IKD to targeted attacks, other perturbation models, and structured prediction tasks.

\bibliographystyle{IEEEtran}
\bibliography{reference}

\end{document}